\documentclass[
10pt, % Main document font size
letterpaper, % Paper type, use 'letterpaper' for US Letter paper, was a4paper
oneside, % One page layout (no page indentation)
headinclude,footinclude, % Extra spacing for the header and footer
]{scrartcl}

\usepackage[
nochapters, % Turn off chapters since this is an article        
beramono, % Use the Bera Mono font for monospaced text (\texttt)
eulermath,% Use the Euler font for mathematics
pdfspacing, % Makes use of pdftex’ letter spacing capabilities via the microtype package
dottedtoc % Dotted lines leading to the page numbers in the table of contents
]{classicthesis} % The layout is based on the Classic Thesis style

\usepackage{arsclassica} % Modifies the Classic Thesis package

\usepackage[T1]{fontenc} % Use 8-bit encoding that has 256 glyphs

\usepackage[utf8]{inputenc} % Required for including letters with accents

\usepackage{graphicx} % Required for including images
\graphicspath{{Figures/}} % Set the default folder for images

\usepackage{enumitem} % Required for manipulating the whitespace between and within lists

\usepackage{lipsum} % Used for inserting dummy 'Lorem ipsum' text into the template

\usepackage{subfig} % Required for creating figures with multiple parts (subfigures)

\usepackage{amsmath,amssymb,amsthm} % For including math equations, theorems, symbols, etc

\usepackage{varioref} % More descriptive referencing

\theoremstyle{definition} % Define theorem styles here based on the definition style (used for definitions and examples)

\theoremstyle{plain} % Define theorem styles here based on the plain style (used for theorems, lemmas, propositions)

\theoremstyle{remark} % Define theorem styles here based on the remark style (used for remarks and notes)

\hypersetup{
colorlinks=true, breaklinks=true, bookmarks=true,bookmarksnumbered,
urlcolor=webbrown, linkcolor=RoyalBlue, citecolor=webgreen, % Link colors
pdftitle={}, % PDF title
pdfauthor={\textcopyright}, % PDF Author
pdfsubject={}, % PDF Subject
pdfkeywords={}, % PDF Keywords
pdfcreator={pdfLaTeX}, % PDF Creator
pdfproducer={LaTeX with hyperref and ClassicThesis} % PDF producer
} % Include the structure.tex file which specified the document structure and layout

\usepackage{graphicx} 
\usepackage{url}
\usepackage{wrapfig}
\usepackage{newtxtext,newtxmath}
\usepackage{tikz}

\makeatletter
\renewcommand*{\@fnsymbol}[1]{\ifcase#1\or*\else\@+\fi}

\title{\normalfont\spacedallcaps{Tractable Boolean and Arithmetic Circuits}\thanks{An earlier version of this article appeared as Chapter~6 in~\cite{NeSyBook}.}} % The article title

\author{\spacedlowsmallcaps{Adnan Darwiche}\thanks{Computer Science Department, University of California, Los Angeles. Email: darwiche@cs.ucla.edu}}

\date{}

\begin{document}

%----------------------------------------------------------------------------------------
%	HEADERS
%----------------------------------------------------------------------------------------

\renewcommand{\sectionmark}[1]{\markright{\spacedlowsmallcaps{#1}}} % The header for all pages (oneside) or for even pages (twoside)
\lehead{\mbox{\llap{\small\thepage\kern1em\color{halfgray} \vline}\color{halfgray}\hspace{0.5em}\rightmark\hfil}} % The header style

\pagestyle{scrheadings} % Enable the headers specified in this block

\maketitle % Print the title/author/date block

\setcounter{tocdepth}{2} % Set the depth of the table of contents to show sections and subsections only

%%%%%%%%%%%%%%%Adnan Macros%%%%%%%%%%%%
%%%%%%%%%%%%%%%%%%%%%%%%%%%%%%%%%%

\newcommand\adnann[1]{\bar{{#1}}}

\def\adnanpr{{\mathit Pr}}

%%PODS

\def\adnanNP{\mathsf{NP}}
\def\adnanPP{\mathsf{PP}}
\def\adnanSP{\mathsf{\#P}}
\def\adnanNPPP{\adnanNP^{\adnanPP}}
\def\adnanPPPP{\adnanPP^{\adnanPP}}

\def\adnansat{{\scshape  Sat}}
\def\adnanms{{\scshape  MajSat}}
\def\adnanems{{\scshape  E-MajSat}}
\def\adnanmms{{\scshape  MajMajSat}}
\def\adnanssat{\#{\scshape  Sat}}

\def\adnanX{{\mathbf  X}}
\def\adnanx{{\mathbf  x}}
\def\adnanY{{\mathbf  Y}}
\def\adnany{{\mathbf  y}}
\def\adnane{{\mathbf  e}}

\def\adnanace{{\scshape  ACE}}
\def\adnanctd{{\scshape  c2d}}
\def\adnanmctd{{\scshape  mini-c2d}}
\def\adnandf{{\scshape  D4}}
\def\adnandsharp{{\scshape  DSHARP}}
\def\adnansdd{{\scshape  SDD}}
\def\adnancudd{{\scshape  CUDD}}
\def\adnanpysdd{{\scshape  PySDD}}
%\def\adnanpypsdd{{\scshape  PyPSDD}}

%%ICML

\def\adnanac{{\mathcal AC}}
\def\adnanmac{{\mathcal MC}}
\def\adnaneql(#1,#2){{#1\!=\!#2}}

%%%%%%%%%%%%%%%%%%%%%%%%%%%%%%%%%%
%%%%%%%%%%%%%%%Adnan Macros%%%%%%%%%%%%

\section*{Abstract} % This section will not appear in the table of contents due to the star (\section*)
Tractable Boolean and arithmetic circuits have been studied extensively in AI for over two decades now.
These circuits were initially proposed as ``compiled objects,'' meant to facilitate logical and probabilistic reasoning,
as they permit various types of inference to be performed in linear-time and a feed-forward fashion like neural networks.
In more recent years, the role of tractable circuits has significantly expanded as they became a 
computational and semantical backbone for some approaches that aim to integrate knowledge, reasoning and learning.
In this article, we review the foundations of tractable circuits and some associated milestones, while focusing on
their core properties and techniques that make them particularly useful for the broad aims of 
neuro-symbolic~AI.

\section{Introduction}
\label{adnan:sec:intro}

Tractable circuits, both Boolean and arithmetic, have been receiving an increased attention in AI and computer
science more broadly. These circuits represent Boolean and real-valued functions, respectively. 
They are called {\em tractable} because they allow one to answer some hard queries about these functions
in polytime, typically through a linear-time, feed-forward pass through the circuit structure. 
The foundations of these circuits have been developed within the area of {\em knowledge
compilation,} which aims to compile knowledge into tractable representations for the goal of facilitating efficient reasoning. 
This area of research has a long tradition in AI; see, e.g.,~\cite{CadoliD97} and~\cite{Marquis95,SelmanK96,DarwicheM02,Darwiche14,DarwicheMSS17,kocoon19FS,DarwichePODS20}. 
A turning point, however, has been the work of~\cite{DarwicheM02} which presented a comprehensive theory
of knowledge compilation based on {\em tractable Boolean circuits,} which are {\em deep} representations,
in contrast to earlier efforts which 
focused on {\em flat} representations based on conjunctive and disjunctive normal forms; 
see, e.g.,~\cite{Marquis95,SelmanK96}. Another turning point has been the work of~\cite{kr/Darwiche02,DarwicheJACM03}
which employed tractable Boolean circuits in probabilistic reasoning, giving birth to an extensive line of
work on {\em tractable arithmetic circuits.} The literature on tractable circuits has enjoyed a number of additional and 
exciting developments over the years, which have broadened and extended their applications from reasoning, 
to learning and more recently to neuro-symbolic AI; an area that is concerned with integrating neural and symbolic
approaches to AI~\cite{Garcez2020NeurosymbolicAT,ijcai/RaedtDMM20,besold2017neuralsymbolic}. 
These developments have also raised key questions, and in some cases
confusions, particularly on tractable arithmetic circuits and their relationship to tractable Boolean circuits and the now
prevalent technique of {\em weighted model counting}~\cite{ai/ChaviraD08}.
The goal of this article is to discuss the foundations of tractable Boolean and arithmetic circuits, to clarify their
relationships, and to highlight some of the key developments that have contributed to the growing interest surrounding 
tractable circuits today. But first, we find it pertinent to further elaborate on how and why the role of tractable
circuits has evolved over the years.

The original motivation behind knowledge compilation and tractable circuits
was based on the following observation. In many reasoning
tasks, one is typically interested in posing a large number of queries so the (high)
cost of {\em offline compilation} can be amortized over the large number of {\em online queries.} 
Today, however, tractable circuits are playing a significantly broader role for a number of reasons. 
First, these circuits have been providing a systematic methodology for computation, particularly for problems 
beyond NP which include important tasks in probabilistic reasoning and machine learning; 
see~\cite{DarwichePODS20} for a recent survey. Second, reasoning with tractable circuits is not only efficient 
but can almost always be conducted using linear-time algorithms that traverse these circuits in a 
feed-forward fashion like neural networks. 
Moreover, tractable circuits are {\em differentiable.} In fact, backpropagation has been performed on these circuits, 
both Boolean~\cite{jancl/Darwiche01} and arithmetic~\cite{DarwicheJACM03}, as early as two decades ago when
the derivatives were first employed in reasoning tasks. 
These properties of tractable circuits made them very suitable for integration
with modern pipelines for machine learning and neuro-symbolic AI; 
see, for example,~\cite{NEURIPS2020_a87c11b9,XuZFLB18,XieXMKS19,aips/LingCK21,pgm/ChenCD20,bnaic/ManhaeveDKDR19,pldi/SaadRM21,tplp/FierensBRSGTJR15,pacmpl/HoltzenBM20,pkdd/DriesKMRBVR15} 
where tractable circuits have been recently employed in and/or integrated with neural networks, deep reinforcement learning,
Bayesian network classifiers and
(deep) probabilistic logic programs.\footnote{Tractable circuits have also attracted interest in areas of
computer science such as database theory; see, e.g.,~\cite{DBLP:journals/mst/JhaS13,kocoon19AA}, and benefited
from areas of computer theory such as communication complexity; see, e.g.,~\cite{DBLP:conf/ijcai/BovaCMS16}.
They also provided tractable approaches for constrained sampling~\cite{SharmaGRM18}, explainable 
AI~\cite{huang2021efficient,corr/quantify-explain,kr/AudemardKM20,arenas2021tractability,broeck2021tractability,kr/ShiSDC20,ijcai/ShihCD18,ecai/DarwicheH20,DarwichePODS20} 
and attracted interest in design tasks such as Boolean functional synthesis~\cite{fmcad/Akshay0CKRS19,lics/ShahBAC21}.
We also remark that (non-tractable) Boolean and arithmetic circuits have their own branches of computational 
complexity theory: circuit complexity and algebraic complexity, respectively. The theory of tractable circuits has
a different focus though compared to what is commonly investigated in these areas.
}
Even the traditional offline/online divide that originally motivated knowledge 
compilation for reasoning is now being exploited in modern settings as it is aligned with the
training/inference divide that governs modern AI systems; see, e.g.,~\cite{NEURIPS2020_a87c11b9}.
As such, a recent trend has emerged in which tools and techniques that were initially
envisioned for reasoning tasks are now being employed to facilitate learning and its integration with
knowledge and reasoning. A few additional milestones have
contributed to broadening the applications of tractable circuits and their expanded role today. 
First is the {\em learning} of tractable arithmetic circuits from data, starting with~\cite{LowdD08}; see also~\cite{ml/RothS09}. 
Second is {\em handcrafting} the structure of these circuits, which started with~\cite{PoonD11}. These 
developments have significantly expanded the utility of tractable circuits as they provided other modes of usage, 
beyond compilation from models.
They also triggered modern treatments of the theory of tractable arithmetic circuits that are
independent of compilation, starting with~\cite{ChoiDarwiche17} that we shall discuss later.
Another milestone is the integration of tractable Boolean and arithmetic circuits, starting with~\cite{KisaBCD14},
which provided a profound, new formalism for integrating symbolic knowledge into
circuits that perform probabilistic reasoning.

This article is organized as follows. We first treat tractable Boolean circuits in Section~\ref{adnan:sec:BC},
followed by tractable arithmetic circuits in Section~\ref{adnan:sec:AC}. We then discuss algorithms that compile knowledge
 into tractable circuits in Section~\ref{adnan:sec:compile} and close with some
remarks in Section~\ref{adnan:sec:conclusion}. 
We do not treat the rich subjects of handcrafting and learning (the structure of) tractable arithmetic circuits
as this is beyond the scope of this article.\footnote{Two video tutorials complement the treatment
in this article: ``Beyond NP with Tractable Circuits,'' \url{https://www.youtube.com/watch?v=kdMzmgyLfQs} 
and ``Three Modern Roles for Logic in AI," \url{https://www.youtube.com/watch?v=3PrYYLppjXA}.}

\section{Tractable Boolean Circuits}
\label{adnan:sec:BC}

\begin{wrapfigure}[6]{r}{0.35\textwidth}
\centering
\vspace{-1mm}
\hspace{0mm}\includegraphics[width=0.35\textwidth]{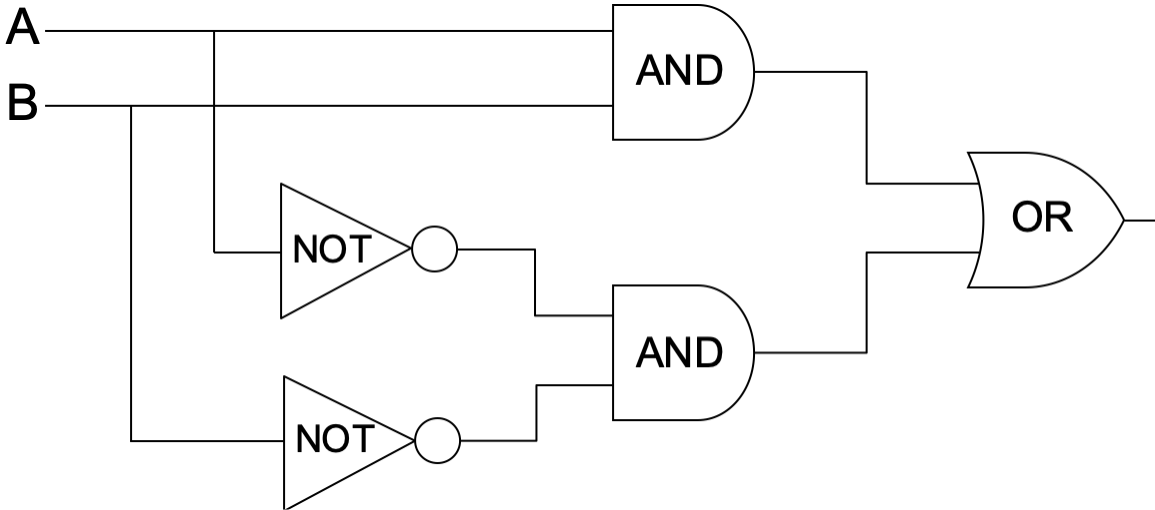}
  \caption*{}
\end{wrapfigure}

The theory of tractable Boolean circuits is based on
{\em Negation Normal Form (NNF)} circuits. These are Boolean circuits that have three types of gates:
and-gates, or-gates and inverters, except that inverters can only feed from the circuit
variables; see figure on the right. 
%Any circuit with these types of gates can be converted to an NNF circuit while at most doubling its size.
NNF circuits are not tractable. However, by imposing certain properties on them we 
can attain different degrees of tractability. A comprehensive, but now incomplete, treatment of tractable 
NNF circuits was given in~\cite{darwicheJAIR02}, in which these circuits were studied
across the two dimensions of {\em tractability} and {\em succinctness.} As we increase the 
strength of properties imposed on NNF circuits, their tractability increases by
 allowing more queries to be performed on them in polytime. This typically
comes at the expense of succinctness as the size of circuits gets larger. We will next review
some classes of tractable Boolean circuits, with increasingly stronger properties, which will allow us 
to efficiently solve decision problems (and their functional variants) that are complete for 
the complexity classes \(\adnanNP \subseteq \adnanPP \subseteq \adnanNPPP \subseteq \adnanPPPP\); see also~\cite{DarwichePODS20}.
The algorithms for these increasingly complex problems all take time linear in
the circuit size,\footnote{The size of a circuit is defined as the number of its edges.} 
and operate by traversing the circuit in a feed-forward fashion like neural networks.
In the following discussion, we will omit inverters from NNF circuits and use \(\neg X\) instead 
to represent an inverted variable \(X\).

\begin{figure}[tb]
  \centering
\includegraphics[width=.9\linewidth]{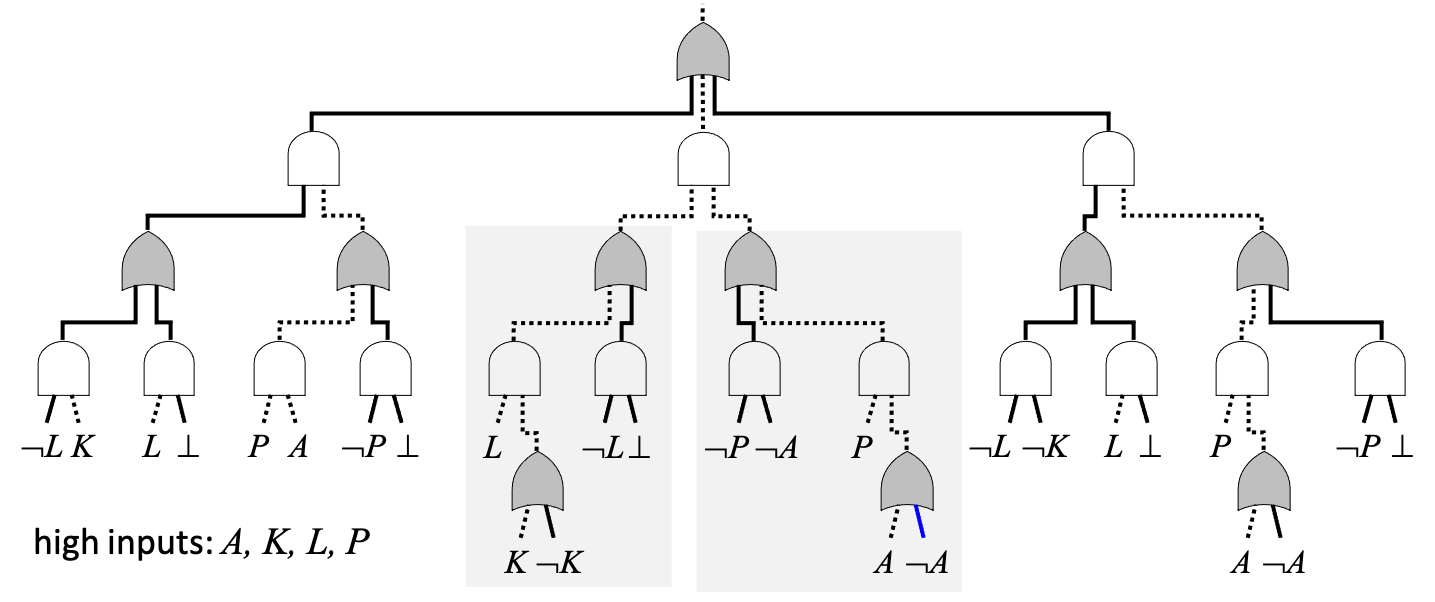}
  \caption{Illustrating the decomposability and determinism properties of NNF circuits. 
  Dotted lines are high wires. Solid lines are low wires.
The illustration does not tie shared inputs of the circuit for clarity of exposition. 
\label{adnan:fig:dnnf}
\label{adnan:fig:d-dnnf}
}
\end{figure}

\smallskip
\noindent {\bfseries Decomposability\ }
One of the simplest properties that make NNF circuits tractable is
{\em decomposability}~\cite{jacm/Darwiche01}. According to this property, 
circuit fragments that feed into an and-gate cannot share variables. 
Figure~\ref{adnan:fig:dnnf} illustrates this property by highlighting two 
fragments (shaded) that feed into an and-gate. The fragment on the left
feeds from variables \(K\) and \(L\). The one on the right
feeds from variables \(A\) and \(P\). NNF circuits that satisfy the 
decomposability property are known as {\em DNNF circuits.}
\adnansat, an $\adnanNP$-complete problem~\cite{satNP}, can be decided in linear time on DNNF 
circuits~\cite{jacm/Darwiche01}. In this context, \adnansat\ is the problem of deciding whether the circuit has 
a satisfying input (generates output \(1\)).

\smallskip
\noindent {\bfseries Determinism\ }
The next property we consider is {\em determinism}~\cite{jancl/Darwiche01}, which applies
to or-gates in an NNF circuit. According to this property, at most one input for an or-gate can
be high under any circuit input. Figure~\ref{adnan:fig:d-dnnf} illustrates this property when 
all circuit variables \(A, K, L, P\) are high. Examining the or-gates in this circuit, under this circuit input,
one sees that each or-gate has either one high input or no high inputs. This property corresponds to
mutual exclusiveness when an or-gate is viewed as a disjunction of its inputs.
NNF circuits that are decomposable and deterministic are known as  {\em d-DNNF circuits}~\cite{jancl/Darwiche01}
and they are exponentially less succinct than DNNF circuits~\cite{DBLP:conf/ijcai/BovaCMS16}.\footnote{That is,
there are Boolean functions that can be represented using DNNF circuits of polynomial size but
their d-DNNF circuits must have exponentially size.}
The $\adnanPP$-complete problem \adnanms~\cite{majSAT} can be decided in polytime on d-DNNF circuits. In this
context, \adnanms\ is the problem of deciding whether the majority of circuit inputs satisfy the circuit. 
If d-DNNF circuits are also {\em smooth}~\cite{jancl/Darwiche01}, a property that can be enforced in quadratic time, 
these circuits allow one to perform \adnanssat~\cite{sharpSAT} in linear time; that is, counting the number 
of satisfying circuit inputs, also known as {\em model counting.} 
If one assigns a weight to each variable value, one can define a weight
for each circuit input as the product of weights assigned to its variable values. One
can then sum the weights of satisfying circuit inputs also in linear time, a problem that is known
as {\em weighted model counting} (WMC)~\cite{ai/ChaviraD08}.

\begin{wrapfigure}[9]{r}{0.35\textwidth}
\centering
%\vspace{-2mm}
\hspace{0mm}  \includegraphics[width=0.35\textwidth]{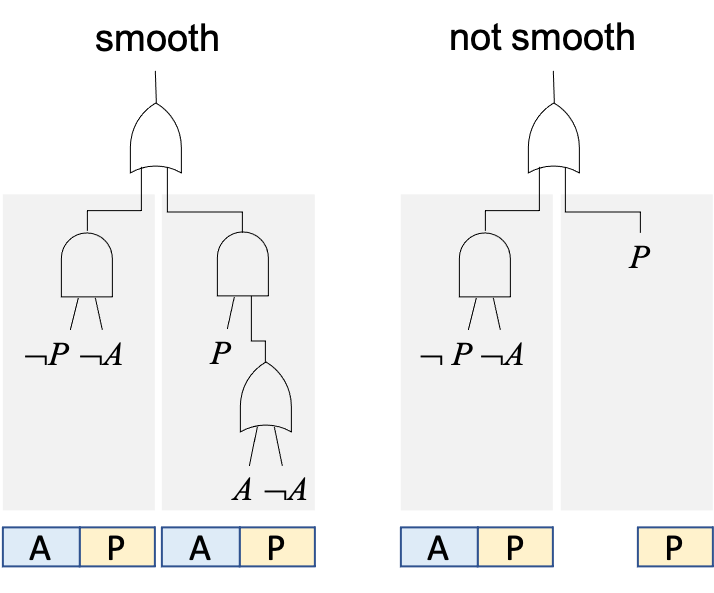}
  \caption*{}
\end{wrapfigure}

\begin{figure}[tb]
\centering
\includegraphics[width=.495\linewidth]{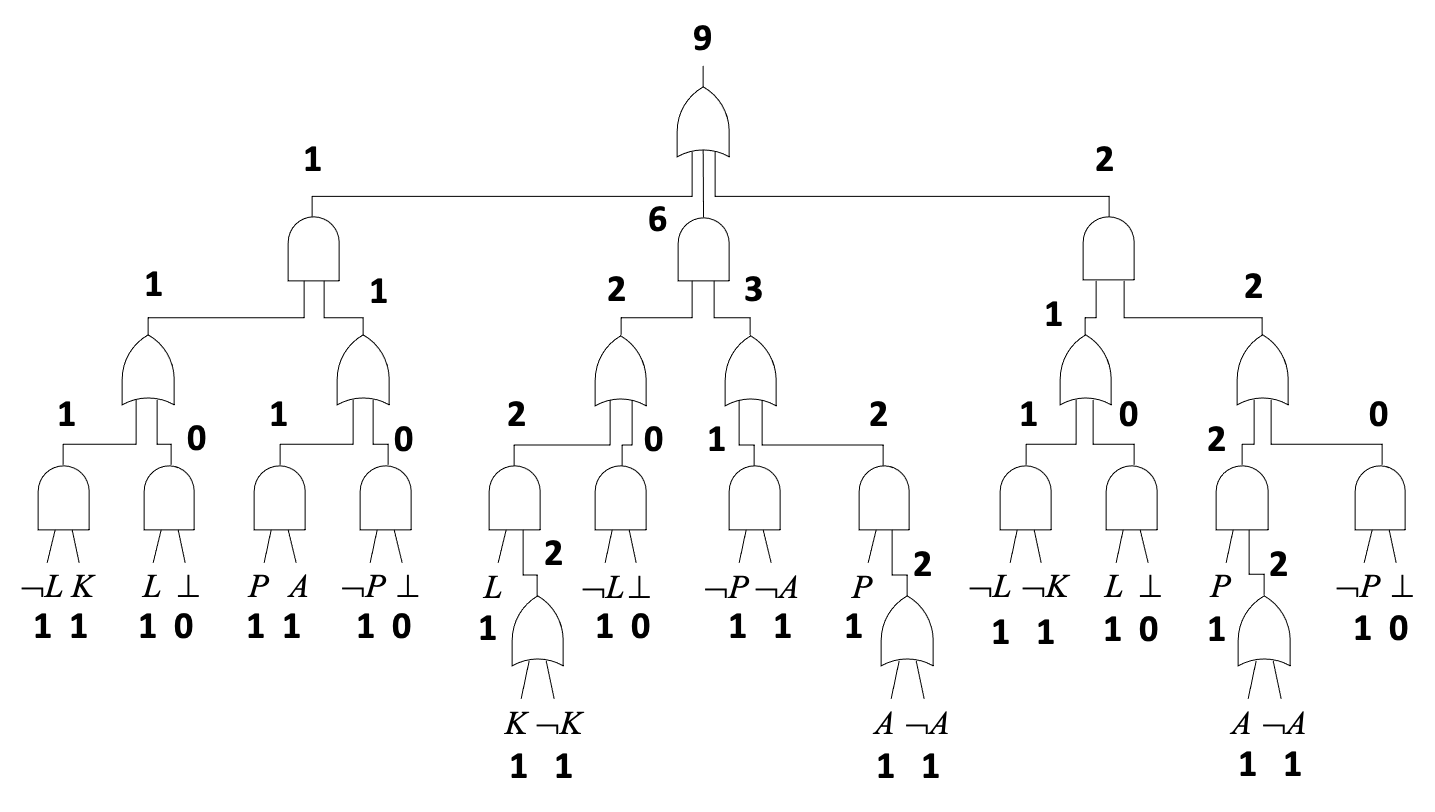}
\includegraphics[width=.495\linewidth]{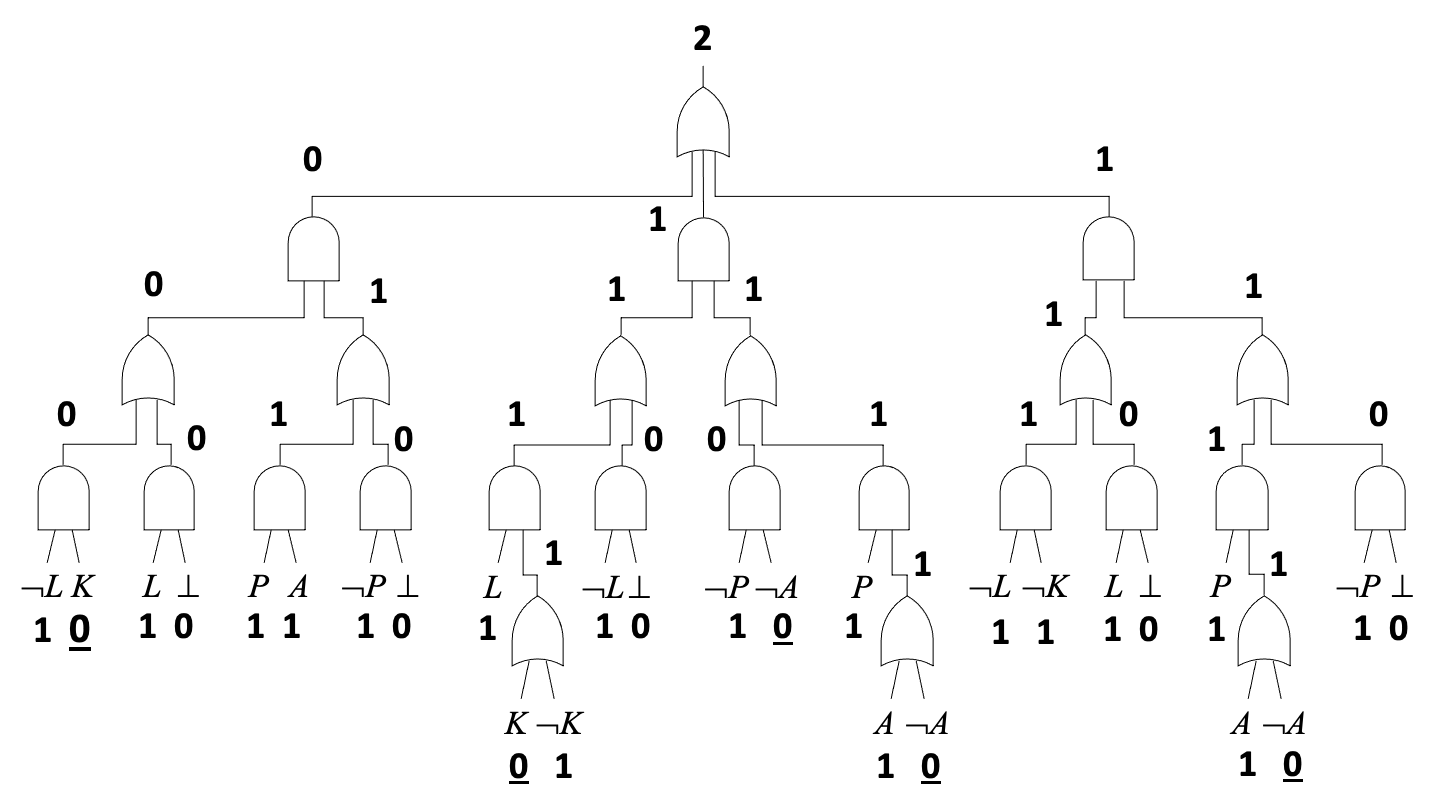}
\caption{Model counting through linear-time circuit traversal. 
Left: counting the number of satisfying circuit inputs. Right: counting the number of satisfying circuit 
inputs with \(\adnaneql(A,1)\) and \(\adnaneql(K,0)\). Unsatisfiable subscircuits always generate a count of \(0\)
even if not smooth, so we can exclude them when checking smoothness.
\label{adnan:fig:mc}
\label{adnan:fig:wmc}}
\end{figure}

\smallskip
\smallskip
\noindent {\bfseries Smoothness\ }
This property requires all circuit fragments feeding into an or-gate to mention the
same variables; see illustration on the right.
Enforcing smoothness can introduce trivial gates
into the circuit, such as the bottom or-gate in the illustration. One can enforce smoothness
in quadratic time~\cite{jancl/Darwiche01} and sometimes more 
efficiently~\cite{DBLP:conf/nips/ShihBBA19}.\footnote{The following 
time-stamped video link provides an intuitive explanation of the role of smoothness in
model counting: \url{https://youtu.be/kdMzmgyLfQs?t=1586}}
An example of model counting using a d-DNNF circuit is depicted in
Figure~\ref{adnan:fig:mc} (left). Every literal in the circuit, whether a positive literal~(\(A\))
or a negative literal~(\(\neg A\)), is assigned the value~\(1\). Constants
\(\top\) and \(\bot\) are assigned the values \(1\) and \(0\), respectively.
We then propagate these numbers upwards, multiplying numbers assigned to the 
inputs of an and-gate and summing numbers assigned to the inputs of an or-gate. 
The number we obtain for the circuit output is the model count. In this example, the circuit has \(9\) satisfying
inputs out of \(16\) possible ones. We can also obtain model counts under variable
settings in linear time. For example, if we wish to count the number of satisfying circuit
inputs in which \(\adnaneql(A,1)\) and \(\adnaneql(K,0)\), we simply assign \(0\) to literals \(\neg A\) 
and \(K\) instead of \(1\) and then propagate counts as just discussed. This is illustrated
in Figure~\ref{adnan:fig:mc} (right), which shows that \(2\) out of the \(9\) satisfying circuit
inputs have \(\adnaneql(A,1)\) and \(\adnaneql(K,0)\). If our goal is to find the model counts
under each setting of a single variable, then all such counts can be obtained using
a second pass on the smooth d-DNNF circuit. This pass
performs backpropagation to compute partial derivatives which can be used to 
obtain these counts~\cite{jancl/Darwiche01}. 
To perform weighted model counting,
we simply assign a weight to a literal instead of the value \(1\) and propagate the
counts as usual.
Model counting is a special case of weighted model counting when each literal weight is \(1\).

\smallskip
\noindent {\bfseries Decision\ }
There are stronger versions of decomposability and determinism which give rise to additional,
tractable NNF circuits.
A stronger version of determinism is known as the {\em decision property.}
It requires each or-gate \(g\) to have the form \(g = (X \wedge \alpha) \vee (\neg X \wedge \beta)\),
where \(X\) is a circuit variable known as the {\em decision variable} of gate \(g\) (this or-gate will
then satisfy the determinism property).
NNF circuits that satisfy decomposability and decision are known as {\em Decision-DNNF}
circuits~\cite{HuangD07} and they are exponentially less succinct than d-DNNF circuits~\cite{jancl/Darwiche01,tods/Beame0RS17}. 
Suppose we split the circuit variables into \(\adnanX\) and \(\adnanY\).
We will say that the Decision-DNNF is {\em \(\adnanX\)-constrained} iff
no or-gate with a decision variable in \(\adnanX\) can appear below an or-gate with a decision
variable in \(\adnanY\)~\cite{PipatsrisawatD09}. These circuits allow us to solve \adnanems\
and its functional variant in time linear in the circuit size.
\adnanems\ is a decision problem that is \(\adnanNPPP\)-complete~\cite{eMajMajSAT}. 
It asks: is there an instantiation \(\adnanx\) under which the majority of instantiations \(\adnany\) yield a satisfying
circuit input \(\adnanx\adnany\)? The functional version of \adnanems\ includes the computation
of most likely partial instantiations in probabilistic reasoning~\cite{ParkD04,PipatsrisawatD09}.
The corresponding feed-forward algorithm 
performs summation at or-gates with decision variables in \(\adnanY\),
maximization at or-gates with decision variables in \(\adnanX\), and multiplication
at and-gates~\cite{PipatsrisawatD09}.
\smallskip
\begin{wrapfigure}[12]{r}{0.15\textwidth}
\centering
%\vspace{-3mm}
%\hspace{-9mm}
  \includegraphics[width=0.15\textwidth]{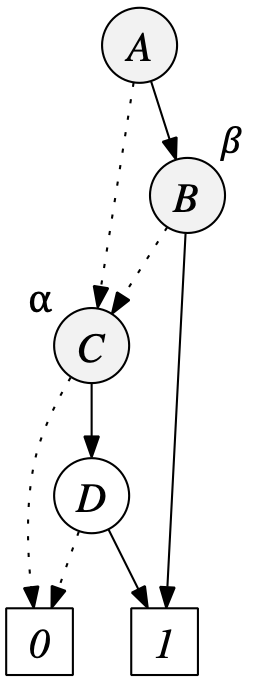}
  \caption*{}
\end{wrapfigure}
{\em Ordered Binary Decision Diagrams (OBDDs)}~\cite{Bryant86} are perhaps one of the most studied,
tractable representations of Boolean functions. OBDDs are a special case of Decision-DNNF circuits
even though they are notated differently as shown on the right. Each internal node in an OBDD is labeled
with a variable and has two outgoing edges: a low edges (usually dotted) and a high edge (usually solid).
The leaf nodes of an OBDD are labeled with \(0\) (\(\bot\)) or \(1\) (\(\top\)). Variables must follow
the same order on any path from the root to a leaf in an OBDD. Consider the root node in the OBDD on
the right, which is labeled with variable \(A\). This node has a low child \(\alpha\) and a high child \(\beta\).
If we replace this node with the or-gate \((\neg A \wedge \alpha) \vee (A \wedge \beta)\) and repeat
the same process recursively for OBDDs \(\alpha\) and \(\beta\), we obtain a Decision-DNNF circuit.
That is, we obtain an NNF circuit that satisfies the decomposability and decision
properties~\cite{DarwicheM02}. 
{\em Free Binary Decision Diagrams (FBDDs)}~\cite{tc/GergovM94} generalize OBDDs by relaxing the variable 
ordering requirement while keeping a weaker property known as {\em test-once:} each variable must appear at most once on any 
path from the root to a leaf. 
FBDDs are also a special case of Decision-DNNF circuits once we adjust for notation and there is a quasipolynomial simulation of 
Decision-DNNF circuits by equivalent FBDDs~\cite{uai/BeameLRS13}. Hence, while FBDDs are less compact than Decision-DNNF circuits,
they are not exponentially less succinct. OBDDs are exponentially less succinct than FBDDs though~\cite{tc/GergovM94}
so they are also exponentially less succinct than Decision-DNNF circuits.

\begin{figure}[tb]
\centering
\begin{tikzpicture}[scale=.95,
                   level 1/.style={sibling distance = .9cm, level distance = 1.1cm},
                   level 2/.style={sibling distance = .45cm, level distance = 1.1cm}] 
\node {\scriptsize 1}
  child{ 
    node {\scriptsize 2}
      child {node {$\scriptsize L$}}
      child {node {$\scriptsize K$}}
  }
  child{
    node {\scriptsize 3}
      child{node {$\scriptsize P$}}
      child{node {$\scriptsize A$}}
  };
\end{tikzpicture}
%%%  }
  \hspace{5mm}
 \begin{tikzpicture}[scale=.95,
                   level 1/.style={sibling distance = 1.4cm, level distance = .8cm},
                   level 2/.style={sibling distance = .6cm, level distance = .8cm}
                   ] 
\node {\scriptsize 1}
  child { 
    node {\scriptsize 2} 
      child {
        node {\scriptsize 3}
          child{node {$A$}}
          child{node { $B$}}
      }
      child{node { $D$}}
  }
  child {
    node {\scriptsize 4} 
    child{node { $C$}}
    child{node { $E$}}
  };
\end{tikzpicture}
%%%  }
  \hspace{5mm}
\begin{tikzpicture}[scale=.95,
                   level 1/.style={sibling distance = .7cm, level distance = 1.1cm},
                   level 2/.style={sibling distance = .7cm, level distance = 1.1cm}] 
\node {\scriptsize 1}
  child {node {$A$}}
  child{
    node {\scriptsize 2}
      child{node {$B$}}
      child{
      node {\scriptsize 3} 
     child{node {$C$}}
     child{node {$D$}}}
  };
\end{tikzpicture}
%%%  }
\caption{Three different types of vtrees: balanced, constrained for \(CE \vert ABD\), and right-linear.
A {\em constrained vtree} for \(\adnanY\vert \adnanX\) is a vtree over variables \(\adnanX \cup \adnanY\) and contains a 
unique node \(u\) with following properties: (1)~\(u\) can be reached from the vtree root by following right children only
and (2)~the variables outside \(u\) are precisely \(\adnanX\). 
A {\em right-linear} vtree is one in which the left child of every internal node is a leaf. 
A right-linear vtree corresponds to a total variable order (\(A, B, C, D, E\) in this case).
\label{adnan:fig:vtree}
\label{adnan:fig:vtrees}}
\end{figure}

\smallskip
\noindent {\bfseries Structured Decomposability\ }
This is a stronger version of decomposability which is stated with respect to a full binary
tree whose leaves are in one-to-one correspondence with the circuit variables~\cite{PipatsrisawatD08}. 
Three such trees are depicted in Figure~\ref{adnan:fig:vtree}, which are known as {\em vtrees.}
Structured decomposability requires each and-gate to have exactly two inputs \(i_1\) and \(i_2\),
and to have a corresponding node \(v\) in the vtree such that the variables of subcircuits feeding into \(i_1\)
and \(i_2\) are in the left and right subtrees of node \(v\). The DNNF 
circuit in Figure~\ref{adnan:fig:dnnf} is structured according to the vtree on the left of Figure~\ref{adnan:fig:vtree}.
For example, all and-gates below the root or-gate conform to vtree node \(v\!=\!1\) in
Figure~\ref{adnan:fig:vtree} (left). 

\begin{figure}[tb]
 \centering
\includegraphics[width=1\linewidth]{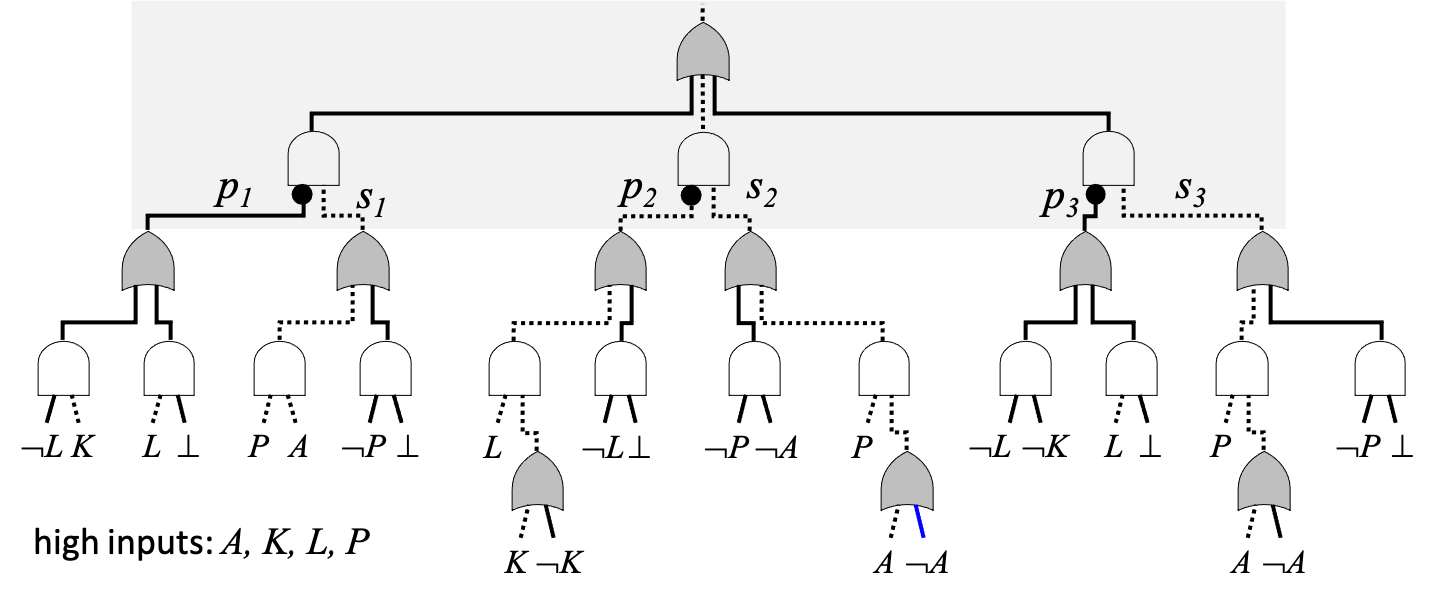}
\caption{Illustrating the partitioned determinism property of NNF circuits. Dotted lines are high wires.
Solid lines are low wires. Precisely one of \(p_1\), \(p_2\) and \(p_3\) will be high for any circuit input.
\label{adnan:fig:sdd}}
\end{figure}

\smallskip
\noindent {\bfseries Partitioned Determinism\ }
Structured decomposability and a stronger version of determinism, known as {\em partitioned determinism,} 
yield a class of NNF circuits known as Sentential Decision Diagrams (SDDs)~\cite{Darwiche11}. These circuits
allow one to solve \adnanmms\ and its functional variant in linear time. \adnanmms\ is a decision problem that
is \(\adnanPPPP\)-complete and is also based on splitting the circuit variables into \(\adnanX\) and \(\adnanY\). It asks: is there a majority of instantiations \(\adnanx\)
under which the majority of instantiations \(\adnany\) yield a satisfying circuit input \(\adnanx\adnany\)? The functional
variant of \adnanmms\ includes the computation of expectations such as~\cite{ijar/ChoiXD12}.
To illustrate partitioned determinism, consider Figure~\ref{adnan:fig:sdd} and the highlighted circuit fragment. 
This fragment corresponds to the Boolean expression \((p_1 \wedge s_1) \vee (p_2 \wedge s_2) \vee (p_3 \wedge s_3)\),
where each \(p_i\) is called a {\em prime} and each \(s_i\) is called a {\em sub} (primes and subs correspond to subcircuits). 
Partitioned determinism requires that
under any circuit input, precisely one prime will be high (i.e., the primes form a partition). In Figure~\ref{adnan:fig:sdd}, 
under the given circuit input,
prime \(p_2\) is high while primes \(p_1\) and \(p_3\) are low. This means that this circuit fragment, which acts
as a multiplexer, is actually passing the value of sub \(s_2\) while suppressing the values of subs \(s_1\) and \(s_3\).
As a result, the or-gate in this circuit fragment is guaranteed to be deterministic: at most one input of the or-gate
will be high under any circuit input. SDD circuits result from recursively applying this multiplexer construct
to a given vtree (the SDD circuit in
Figure~\ref{adnan:fig:sdd} is structured with respect to the vtree on the left of Figure~\ref{adnan:fig:vtree}).\footnote{The vtree 
of an SDD is ordered: the distinction between left and right children matters.}
Recall that \adnanmms, the prototypical 
problem for the complexity class $\adnanPPPP$~\cite{eMajMajSAT}, is stated with respect to a split of circuit 
variables into \(\adnanX\) and \(\adnanY\). If the vtree is {\em constrained for \(\adnanY|\adnanX\),} then this problem and its functional variant can be 
solved in linear time on the corresponding SDD~\cite{kr/OztokCD16}.
Figure~\ref{adnan:fig:vtrees} illustrates the concept of a constrained vtree.
When an SDD is structured with respect to a {\em right-linear vtree,} 
the result corresponds to an OBDD after adjusting for notation as discussed earlier.
Figure~\ref{adnan:fig:vtrees} illustrates the concept of a right-linear vtree.
In this case, every circuit fragment will have
the form \((X \wedge \alpha) \vee (\neg X \wedge \beta)\) where literals \(X\) and \(\neg X\) are
primes, and where \(\alpha\) and \(\beta\) are subs.
OBDDs are therefore a subset of both FBDDs and SDDs and they are exponentially less 
succinct than SDDs~\cite{aaai/Bova16}. 
SDDs and FBDDs are not comparable though in terms of succinctness~\cite{uai/BeameL15,mst/BolligB19}.\footnote{SDDs
can be exponentially less succinct than FBDDs for some Boolean functions~\cite{uai/BeameL15} and FBDDs
can be exponentially less succinct than SDDs for some other functions~\cite{mst/BolligB19} so these circuit types
are not comparable. Hence, SDDs and Decision-DNNFs are not comparable either since FBDDs
are a subset of Decision-DNNFs and can simulate them quasipolynomially~\cite{uai/BeameLRS13}.
However, both SDDs and Decision-DNNFs are exponentially less succinct than d-DNNF circuits
since FBDDs are exponentially less succinct than d-DNNF circuits~\cite{jancl/Darwiche01}.}

We have covered in this section only a subset of the tractable Boolean circuits known today, but ones that provide
the basis for many of the further refinements and additions; see, e.g,~\cite{aaai/LaiMY21}
and~\cite{lics/ShahBAC21} for some recent additions and~\cite{kocoon19FS} for a relatively recent tutorial
that covers more circuit types and discusses further the relative succinctness of these circuits.
The circuit properties we covered also form a basis
for the most influential types of tractable arithmetic circuits. We shall cover these circuit types in the next section.

\section{Tractable Arithmetic Circuits}
\label{adnan:sec:AC}

\begin{figure}[tb]
\centering
\includegraphics[width=\linewidth]{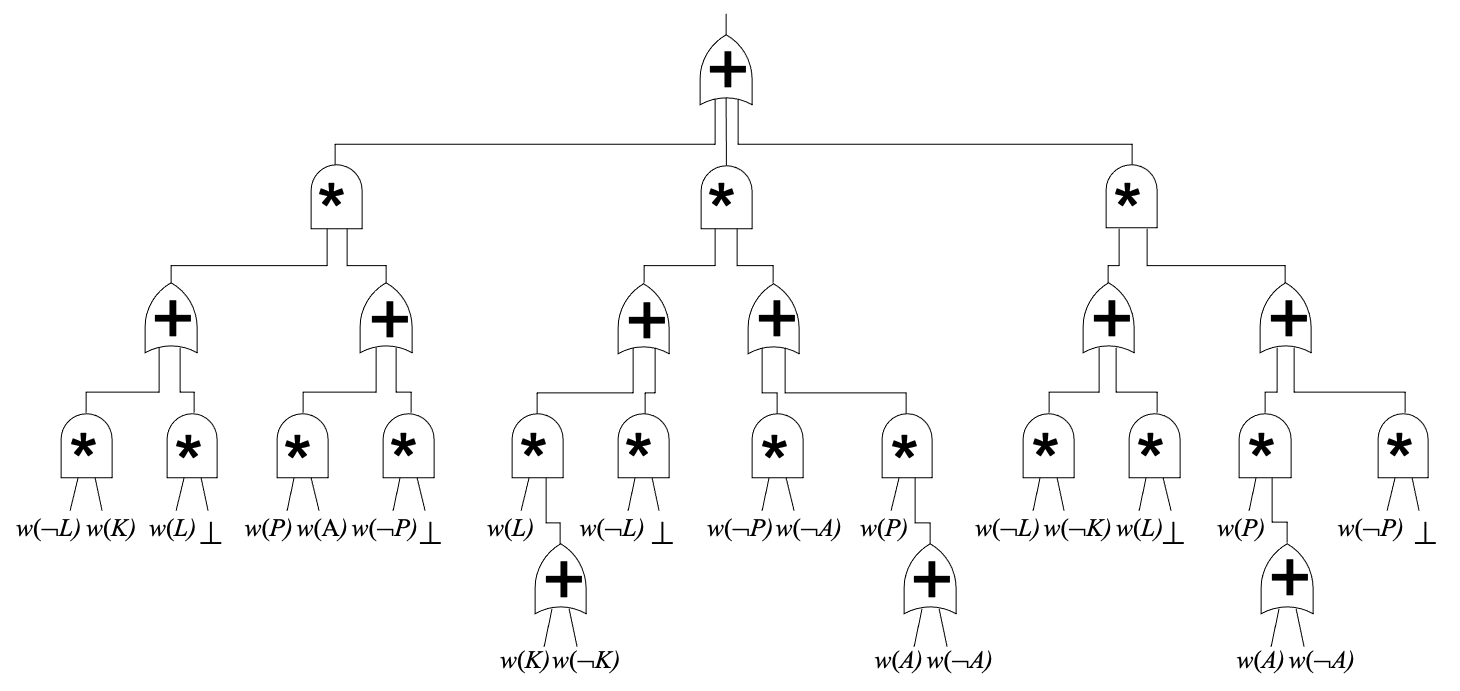}
%\quad
%\includegraphics[width=.48\linewidth]{figures/BC2maxAC}
\caption{Computing the weighted model count of a tractable Boolean circuit, where \(w(\ell)\) 
represents the weight of literal \(\ell\). This computation induces an arithmetic circuit that is
superimposed on the Boolean circuit.
\label{adnan:fig:BC2AC}}
\end{figure}

We saw in the previous section how counting the models of a tractable Boolean circuit is done through the application of
arithmetic operations at Boolean gates: additions at or-gates and multiplications at and-gates. 
This counting task induces an arithmetic circuit that shares
the structure of underlying Boolean circuit as shown in Figure~\ref{adnan:fig:BC2AC},
therefore inheriting its properties such as decomposability, determinism and smoothness.
Decomposability and smoothness, being structural properties, maintain their exact definitions when moving from the
Boolean to the arithmetic side. However, determinism ends up being phrased slightly differently as we shall see later. 

\begin{figure}[tb]
\centering
\includegraphics[width=\linewidth]{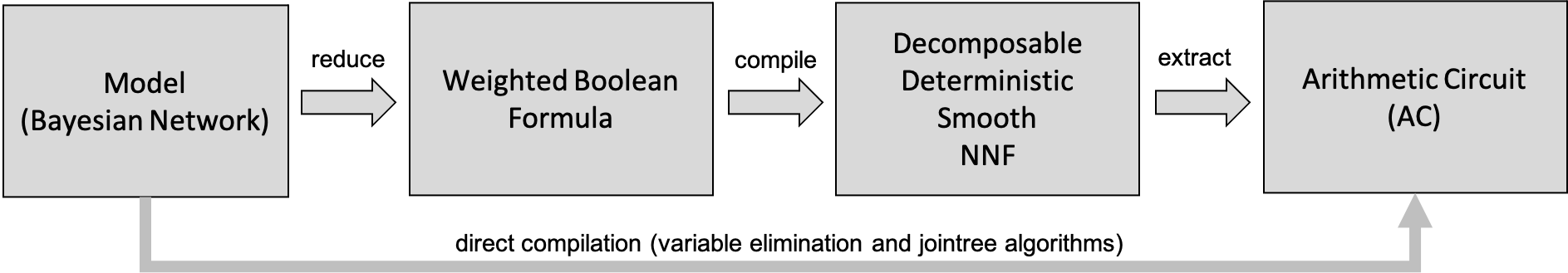}
\caption{Constructing arithmetic circuits through a compilation process.
\label{adnan:fig:pipeline}}
\end{figure}

The above connection between tractable Boolean and arithmetic circuits originated from~\cite{kr/Darwiche02,DarwicheJACM03},
which compiled Bayesian networks into tractable arithmetic circuits to enable linear-time probabilistic 
reasoning on the compiled circuits; see Figure~\ref{adnan:fig:pipeline}.
According to this proposal, the Bayesian network is first encoded into a Boolean formula with literal weights,
allowing one to reduce probabilistic reasoning into weighted model counting. 
The Boolean formula is then compiled into a tractable Boolean circuit (deterministic, decomposable and smooth), 
from which a tractable arithmetic circuit known as an AC is finally extracted. 
This approach is implemented by the \adnanace\ system,\footnote{\url{http://reasoning.cs.ucla.edu/ace/}} 
which was recently evaluated in~\cite{KuldeepIJCAI21} and shown to exhibit state-of-the-art performance; see also~\cite{uai/DilkasB21}.

A more direct and modern account of tractable arithmetic circuits is given in~\cite{ChoiDarwiche17},
which reconstructed the proposal in~\cite{DarwicheJACM03} so it is independent of Bayesian networks, 
model compilation and weighted model counting. This modern treatment was motivated by two influential lines of 
developments. The first line, which started with~\cite{LowdD08}, utilized arithmetic circuits in the context of learning,
in contrast to reasoning. 
The second line of developments, initiated by~\cite{PoonD11}, handcrafted the structure of arithmetic circuits 
instead of compiling them from models and dropped the property of determinism while keeping the circuits
tractable for certain computations. 
These developments raised key questions but also broadened the applications of tractable arithmetic circuits significantly. 
We will next present the modern treatment of~\cite{ChoiDarwiche17} which will allow us to also explain
more recent classes of tractable arithmetic circuits, such
as Sum-Product Networks (SPNs)~\cite{PoonD11} and Probabilistic Sentential Decision Diagrams (PSDDs)~\cite{KisaBCD14}. 
As we shall see, the treatment in~\cite{ChoiDarwiche17} is based on a key
distinction between arithmetic circuits, which can lookup values, and tractable arithmetic circuits,
which can also reason. As in the Boolean case, the degree to which an arithmetic circuit is tractable and, hence,
its ability to conduct various forms of reasoning, depends on the specific properties that the circuit satisfies.

\subsection*{The Reference Point of an Arithmetic Circuit}

\begin{figure}[tb]
\centering
%%\subfigure[\(f_1(A)\)]{\label{adnan:fig:factor1}
\small
\begin{tabular}[b]{c|c}
\(A\) & \(f_1\) \\\hline
$a$ & $1$ \\
$\adnann{a}$ & $2$
\end{tabular}
%%}
\quad \quad
%%\subfigure[\(f_2(A,B)\)]{\label{adnan:fig:factor2}
\small
\begin{tabular}[b]{cc|c}
\(A\) & \(B\) & \(f_2\) \\\hline
$a$ & $b$ & $3$ \\
$a$ & $\adnann{b}$ & $4$ \\
$\adnann{a}$ & $b$ & $5$ \\
$\adnann{a}$ & $\adnann{b}$ & $6$ 
\end{tabular}
%%}
\quad \quad
%%\subfigure[\(f = f_1 f_2\)]{\label{adnan:fig:factor3}
\small
\begin{tabular}[b]{cc|c}
\(A\) & \(B\) & \(f\) \\\hline
$a$ & $b$ & $3$ \\
$a$ & $\adnann{b}$ & $4$ \\
$\adnann{a}$ & $b$ & $10$ \\
$\adnann{a}$ & $\adnann{b}$ & $12$
\end{tabular}
\includegraphics[width=.45\linewidth]{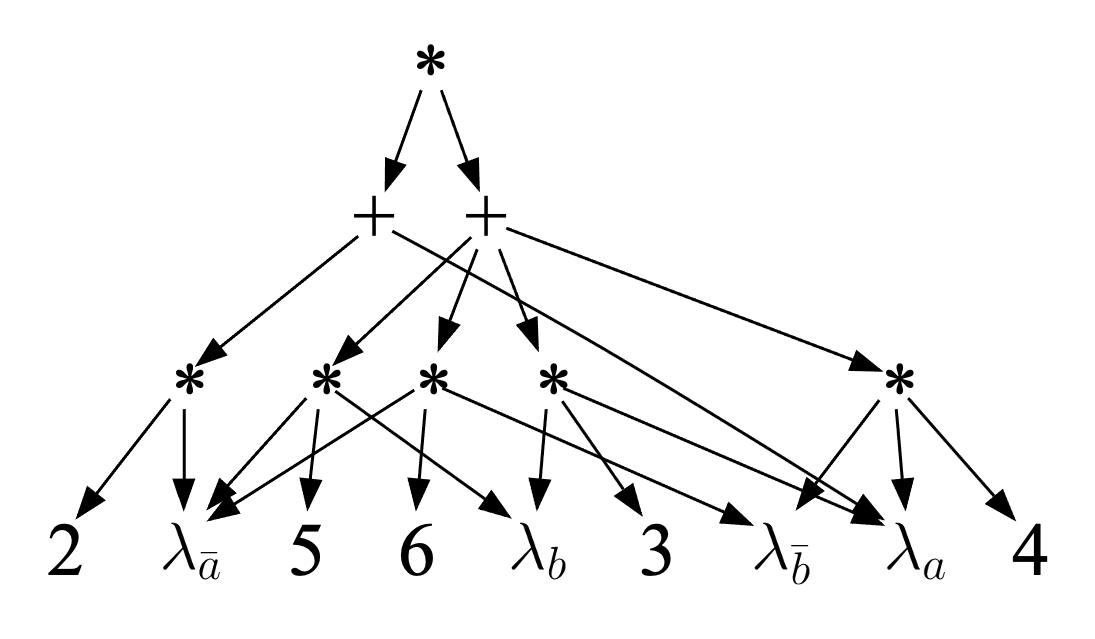}
%%}
\caption{Factors \(f_1(A)\), \(f_2(A,B)\), their product factor \(f(A,B) = f_1 f_2\), and
an arithmetic circuit \(\adnanac_1\) that computes factor \(f(A,B)\).
The circuit indicators (\(\lambda_a,\lambda_{\adnann{a}},\lambda_b,\lambda_{\adnann{b}}\)) correspond 
to the values of variables \(A\) and \(B\).
The variables of factors and circuits can be discrete in general (they are binary in this example).
\label{adnan:fig:factors}\label{adnan:fig:factor1}\label{adnan:fig:factor2}\label{adnan:fig:factor3} \label{adnan:fig:AC1-def}
}
\end{figure}

There is a key difference between arithmetic circuits that are compiled from models and those 
that are handcrafted or learned from data. The former have a clear reference point, the model,
which defines the quantities that these circuits are supposed to compute. The latter circuits lack such a 
reference point which can cause issues when stating and evaluating claims. 
The theory in~\cite{ChoiDarwiche17} starts by defining the reference point of an arithmetic
circuit, regardless of where the circuit originates from. This reference point is the notion of a {\em factor:} 
a generalization of a Boolean function that maps {\em complete} variable instantiations into non-negative 
numbers; see Figure~\ref{adnan:fig:factors} (a distribution is one type of a factor).

\begin{figure}[tb]
\centering
\includegraphics[width=.45\linewidth]{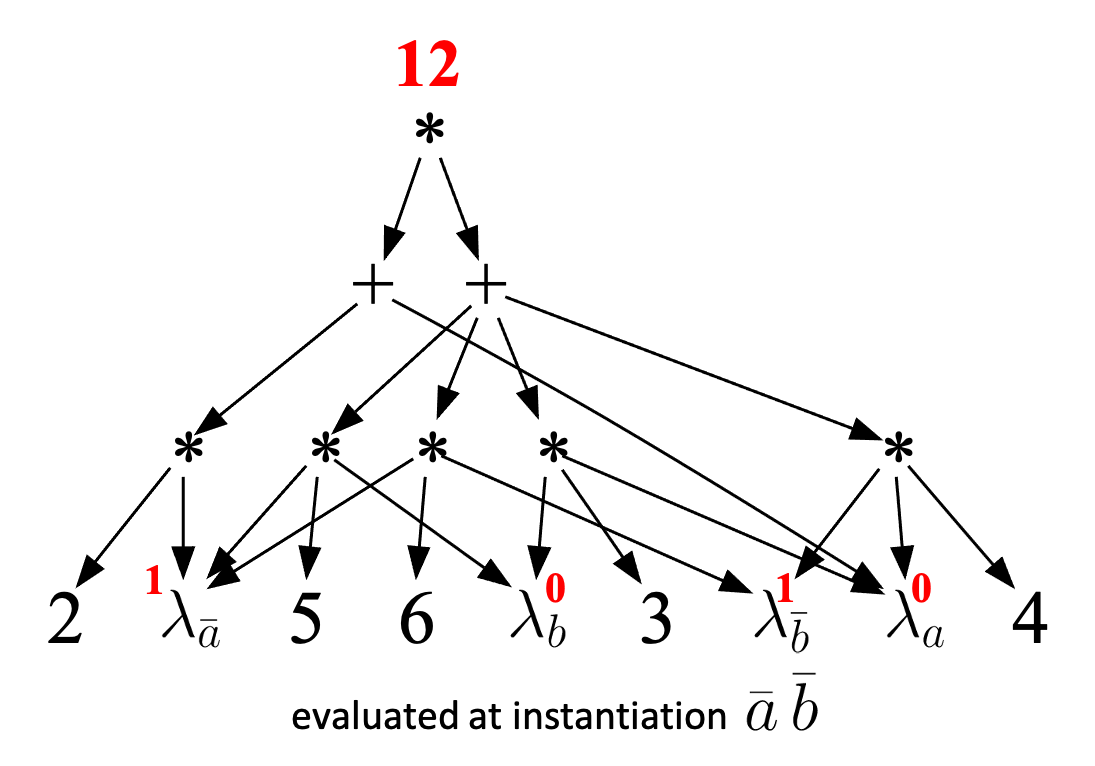}
\quad
\includegraphics[width=.45\linewidth]{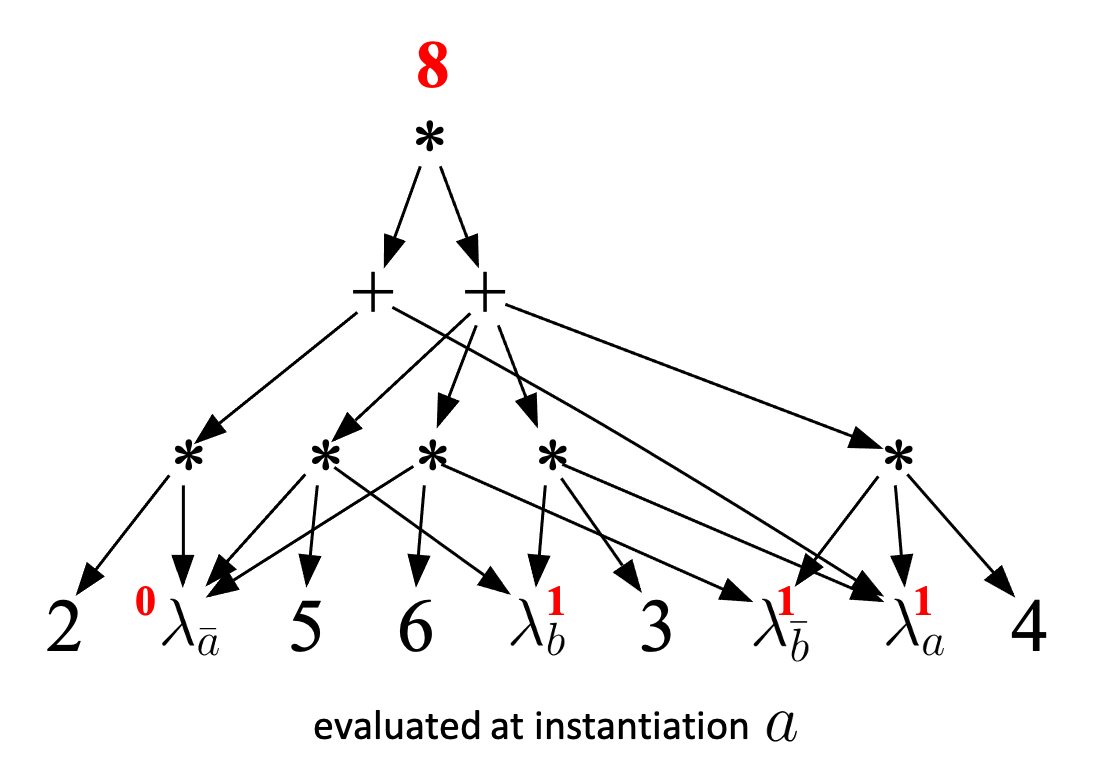}
\caption{An arithmetic circuit \(\adnanac_1\) evaluated at complete variable instantiation \(\adnaneql(A,\adnann{a}),\adnaneql(B,\adnann{b})\) 
and partial variable instantiation \(\adnaneql(A,a)\). 
This circuit computes factor \(f(A,B)\) in Figure~\ref{adnan:fig:factors}: it can lookup the values of factor \(f\).
This circuit is {\bfseries not decomposable} but is {\bfseries smooth} and {\bfseries deterministic.}
\label{adnan:fig:AC1-eval}}
\end{figure}

An arithmetic circuit is based on a set of {\em discrete variables,} which define a key ingredient of 
the circuit: the {\em indicators.} For each value \(x\) of a variable \(X\), we have an indicator \(\lambda_x\). 
The arithmetic circuit will then have constants
and indicators as its leaf nodes (inputs) with adders and multipliers as its internal nodes; see Figure~\ref{adnan:fig:AC1-def}.
The factor of an arithmetic circuit (the reference point) is obtained by evaluating the circuit at complete variable instantiations.
To evaluate the circuit at an instantiation \(\adnane\), we replace each indicator \(\lambda_x\) 
with \(1\) if the value \(x\) is compatible with instantiation \(\adnane\) and with \(0\) otherwise~\cite{DarwicheJACM03}. 
We then evaluate the circuit bottom-up in the standard way.
The factor \(f(A,B)\) in Figure~\ref{adnan:fig:AC1-def} has four rows which correspond to the four instantiations of variables \(A\) and \(B\). 
Evaluating the circuit in this figure at each of these complete instantiations 
yields a value for each instantiation and therefore defines the reference factor.
We say in this case that the arithmetic circuit {\em computes} this factor. We also say that this circuit
can {\em lookup} the values of this factor. 
For example, in Figure~\ref{adnan:fig:AC1-eval} (left), the circuit evaluates to \(12\) under the complete variable
instantiation \(\adnaneql(A,{\adnann{a}}),\adnaneql(B,{\adnann{b}})\) by setting the indicators to
\(\adnaneql({\lambda_a},0), \adnaneql({\lambda_{\adnann{a}}},1), \adnaneql({\lambda_b},0), \adnaneql({\lambda_{\adnann{b}}},1)\). 
An arithmetic circuit can be evaluated at a {\em partial} variable 
instantiation using the same procedure, but the value returned may not be meaningful unless the circuit is 
tractable (i.e., unless the circuit satisfies certain properties). 
For example, Figure~\ref{adnan:fig:AC1-eval} (right) evaluates the circuit at partial instantiation \(\adnaneql(A,a)\)
by setting the indicators to
 \(\adnaneql({\lambda_a},1), \adnaneql({\lambda_{\adnann a}},0), \adnaneql({\lambda_b},1), \adnaneql({\lambda_{\adnann b}},1)\), 
 leading to a value of \(8\). 
However, this value is not meaningful since the circuit is not tractable so it cannot reason about the factor. 
This is a subject that we shall discuss in depth next.
%\footnote{The values of indicators do not have to be \(0\) or \(1\). 
%For example, if the factor computed by a circuit represents a distribution, and if the circuit satisfies certain properties,
%then we can use the circuit to reason about uncertain evidence by setting the indicators to real values. 
%More on this in footnote~\ref{adnanfoot:soft-evidence}.}

\subsection*{Circuits that Lookup Values Versus Circuits that Reason}

The central question posed and treated in~\cite{ChoiDarwiche17} is the following. 
Suppose we have an arithmetic circuit that {\em computes} a factor (i.e., looks up its values).
Can this circuit {\em reason} about the factor and why?
Constructing an arithmetic circuit that computes a factor can be
done efficiently even when the factor is defined implicitly as a product of other factors (e.g., when 
the factor is defined by a probabilistic graphical model). Consider the factors and arithmetic circuit in 
Figure~\ref{adnan:fig:factors}. Factor \(f(A,B)\) is the product of factors \(f_1(A)\) and \(f_2(A,B)\). 
The arithmetic circuit computes factor \(f(A,B)\) and is obtained by multiplying circuits
\(\lambda_{a} + 2\lambda_{\bar{a}}\) and \(3\lambda_{a} \lambda_{b} + 
4\lambda_{a} \lambda_{\bar{b}} + 
5\lambda_{\bar{a}} \lambda_{b} + 
6\lambda_{\bar{a}} \lambda_{\bar{b}}\), which compute factors \(f_1(A)\) and \(f_2(A,B)\), respectively.
This can always be done and the size of resulting circuit is linear in the size of multiplied
factors, not the size of their product which can be exponential.\footnote{To see this,  
define the depth-two arithmetic circuit of factor \(f_i(\adnanX)\) as follows:
\(
\sum_\adnanx f_i(\adnanx) \prod_{x \sim \adnanx} \lambda_x,
\)
where \(x \sim \adnanx\) means that value \(x\) of variable \(X \in \adnanX\) is compatible with 
instantiation \(\adnanx\) of variables \(\adnanX\). This arithmetic
circuit has one layer of multipliers followed by a layer with one adder, and computes factor \(f_i(\adnanX)\). 
If a factor \(f\) is defined as the product of
factors \(f_1, \ldots, f_n\), then multiplying the depth-two circuits of factors \(f_i\) yields a circuit that
computes their product factor \(f = f_1, \ldots, f_n\); see~\cite{ChoiDarwiche17} for details.
}

The interest, however, is in {\em tractable} arithmetic circuits that can reason about a factor, not ones that can only look up its 
values. One fundamental reasoning task is that of computing the value of a partial variable 
instantiation, known as the {\em marginals (MAR)} problem~\cite{Pearl88b}. 
Another fundamental reasoning task is that of identifying and computing the value of
a {\em most likely, complete variable instantiation,} known as the MPE problem~\cite{Pearl88b}.
For factors that are induced by models such as Bayesian networks, the decision variant of MPE
is \(\mathrm{NP}\)-complete~\cite{Shimony94}, the decision variant of MAR 
is \(\mathrm{PP}\)-complete and its functional variant is \(\adnanSP\)-complete~\cite{Roth96}, making these hard reasoning
tasks.\footnote{MPE stands for {\em Most Likely Explanation.}
A related problem is that of identifying and computing
the value of a most likely instantiation of a partial set of variables, which is known as the MAP problem~\cite{Pearl88b}.
The decision variant of this problem is \(\mathrm{NP}^{\mathrm{PP}}\)-complete~\cite{ParkD04}; 
see also \cite{Darwiche09}. MAP stands for {\em Maximum a Posteriori Hypothesis.} 
Sometimes, MAP and partial MAP are used instead of MPE and MAP to denote these problems.}

Consider again the factor \(f(A,B)\) in Figure~\ref{adnan:fig:AC1-def} and the partial instantiation \(\adnaneql(A,a)\).
The marginal for this instantiation, \(f(a)\), is the sum of values assigned to rows that are compatible with 
instantiation \(\adnaneql(A,a)\):
\(f(a) = f(a,b)+f(a,\adnann{b}) = 3+4 = 7\). This computation is quite fundamental as it 
corresponds to computing marginal probabilities when the factor represents a distribution.
Interestingly enough, while the arithmetic circuit \(\adnanac_1\) of Figure~\ref{adnan:fig:AC1-def} does compute the factor, 
it does not compute its marginals. That is, it can lookup values of rows but cannot sum them up (i.e., cannot reason).
A counterexample is shown on the right of Figure~\ref{adnan:fig:AC1-eval}, 
where the circuit evaluates to \(8\) instead of \(7\) at instantiation \(\adnaneql(A,a)\). 

\begin{figure}[tb]
\centering
\includegraphics[width=.45\linewidth]{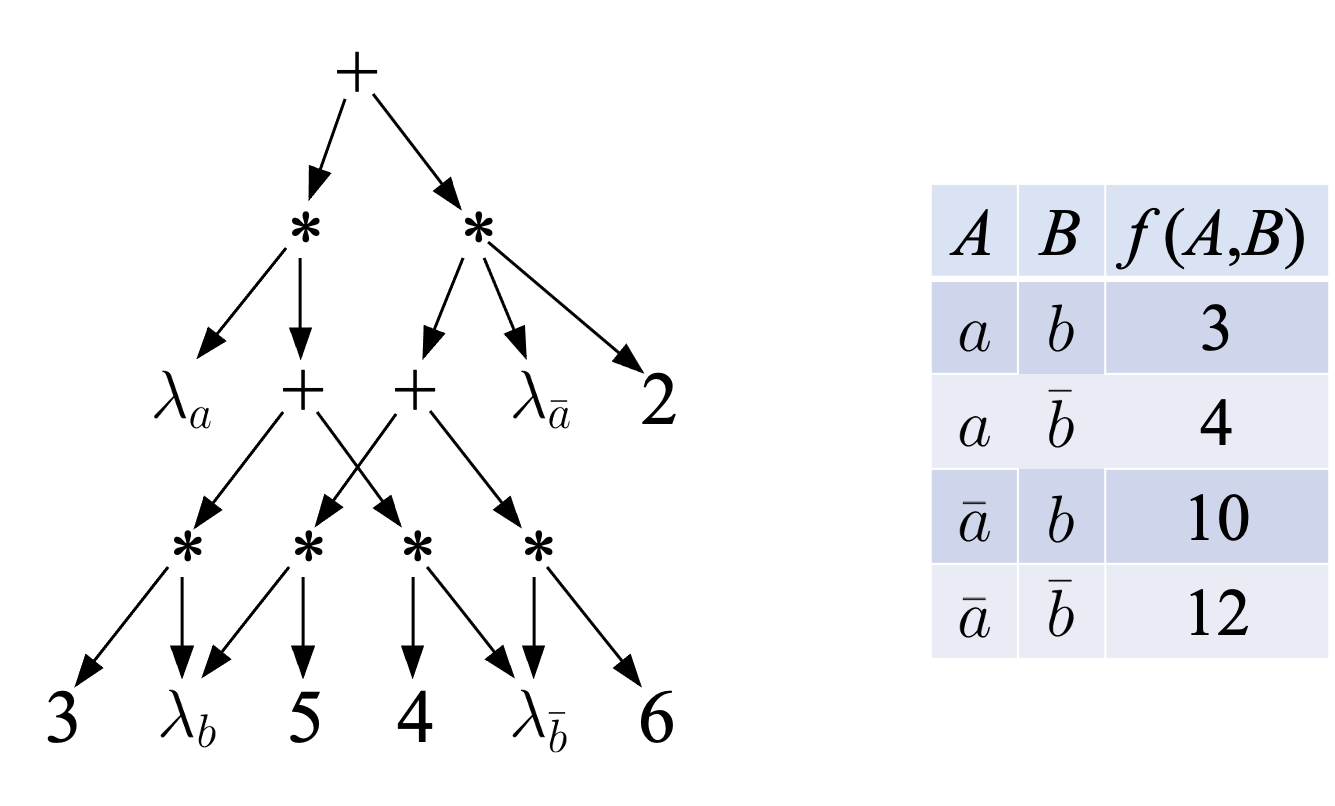}
\quad
\includegraphics[width=.23\linewidth]{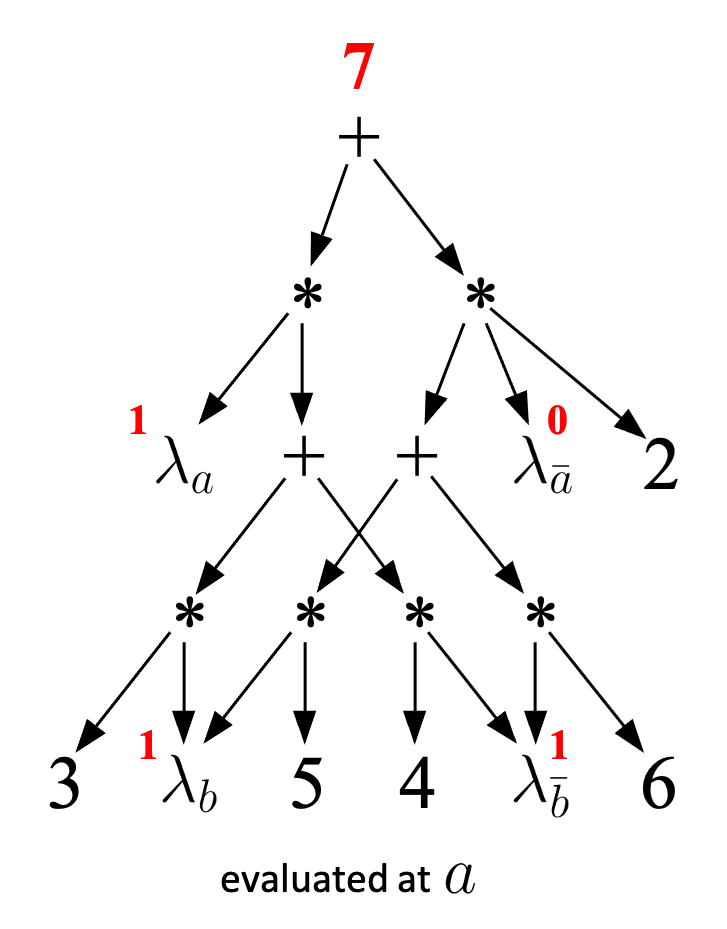}
\quad
\includegraphics[width=.23\linewidth]{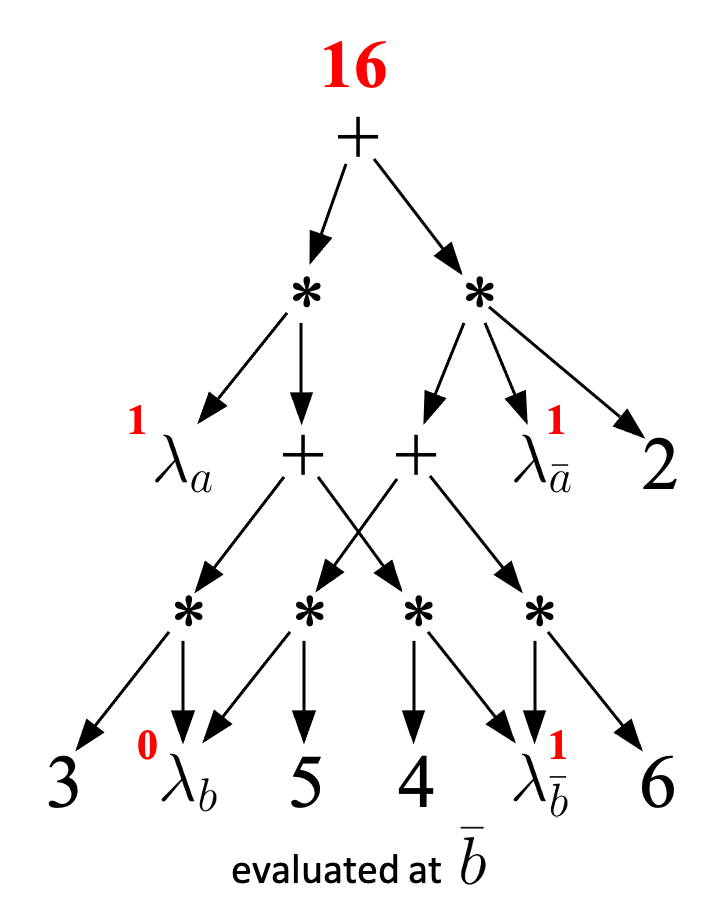}
\caption{An arithmetic circuit \(\adnanac_2\), the factor it computes, and evaluations of the circuit
under the partial variable instantiations \(\adnaneql(A,a)\) and \(\adnaneql(B,\adnann{b})\).
This arithmetic circuit computes the factor marginals.
\label{adnan:fig:AC2-def}\label{adnan:fig:AC2-eval}}
\end{figure}

Figure~\ref{adnan:fig:AC2-def} depicts another arithmetic circuit, \(\adnanac_2\), which computes the same factor 
computed by \(\adnanac_1\) of Figure~\ref{adnan:fig:AC1-def}. Unlike \(\adnanac_1\) though, \(\adnanac_2\) does compute the 
factor marginals.
Figure~\ref{adnan:fig:AC2-eval} depicts example evaluations of \(\adnanac_2\) to be 
contrasted with the evaluations of \(\adnanac_1\) in Figure~\ref{adnan:fig:AC1-eval}. The question now is:
Why did \(\adnanac_2\) compute marginals, and hence reason about the factor, while \(\adnanac_1\)
could not? Before we answer this question, let us first examine another example of reasoning
about a factor:
computing the value and identity of a most likely, complete variable instantiation (MPE).

\begin{figure}[tb]
\centering
\includegraphics[width=.7\linewidth]{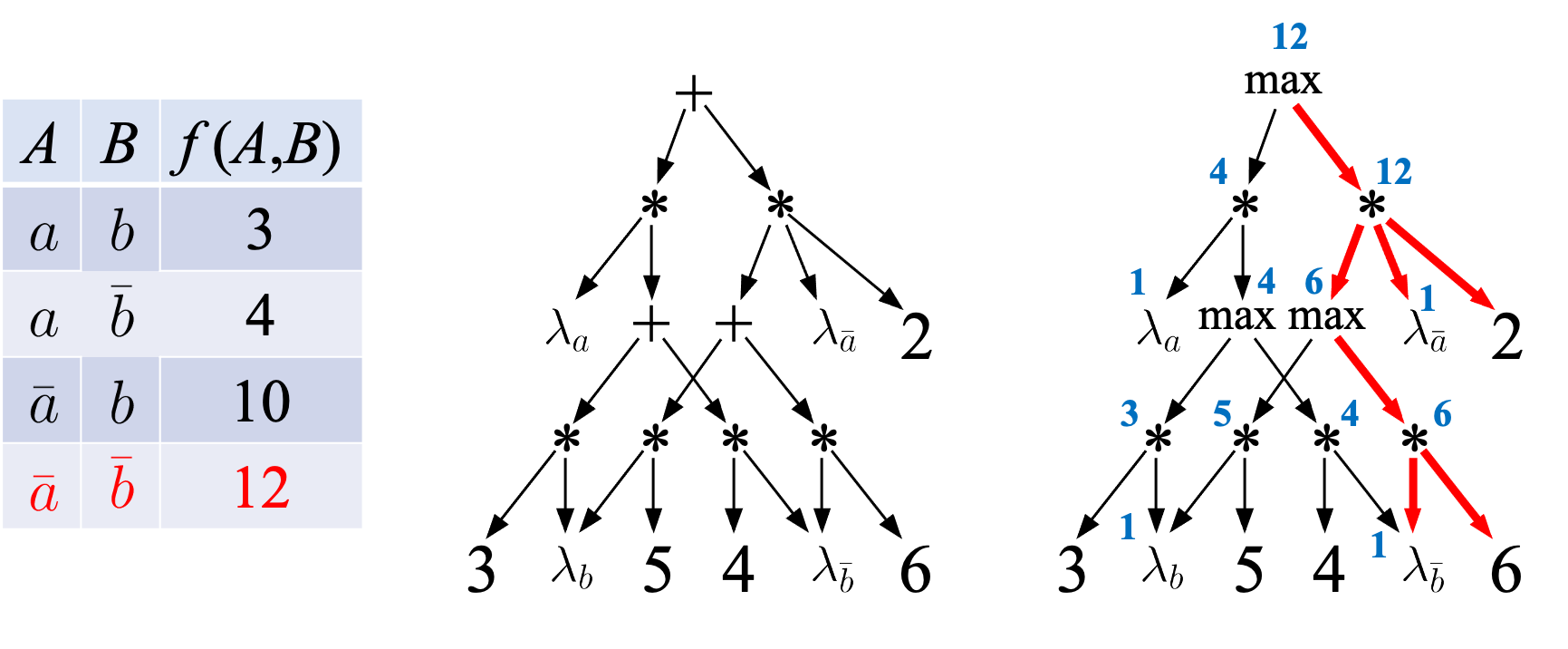}
\caption{A factor, an arithmetic circuit \(\adnanac_2\) that computes this factor, and the corresponding 
maximizer circuit \(\adnanmac_2\).  
This maximizer circuit computes the factor MPEs.
\label{adnan:fig:AC2-MPE}}
\end{figure}

Consider Figure~\ref{adnan:fig:AC2-MPE} which depicts \(\adnanac_2\) again and the factor it computes.
The figure shows another circuit, \(\adnanmac_2\), obtained by replacing the adders of \(\adnanac_2\) with maximizers.
This is called a {\em maximizer arithmetic circuit,} introduced in~\cite{ChanD06}. If we evaluate this circuit at the
empty variable instantiation by setting all indicators to \(1\), we obtain a value of
\(12\) which is the maximal value attained by any row of the factor; see the right of
Figure~\ref{adnan:fig:AC2-MPE}. If we further evaluate this maximizer circuit at instantiation
\(\adnaneql(B,b)\), we obtain a value of \(10\) which is the maximal value attained by any row of the factor
that is compatible with instantiation \(\adnaneql(B,b)\). One can verify that this maximizer
circuit will compute the factor MPEs correctly for any variable instantiation, whether complete or partial. 

\begin{figure}[tb]
\centering
\includegraphics[width=\linewidth]{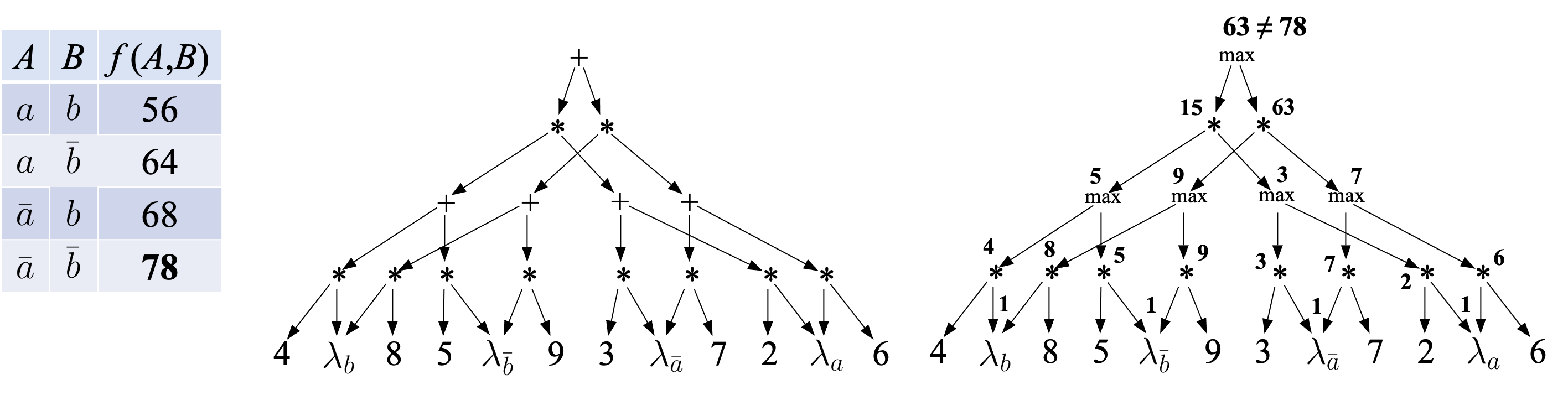}
\caption{A factor, an arithmetic circuit \(\adnanac_3\) that computes the factor marginals, 
and a maximizer circuit \(\adnanmac_3\) that does not compute the factor MPEs.
\label{adnan:fig:AC3-no-MPE}}
\end{figure}

Consider now Figure~\ref{adnan:fig:AC3-no-MPE} which depicts a third arithmetic circuit, \(\adnanac_3\), and the factor
it computes. This circuit computes the factor marginals but does not compute the factor MPEs. The figure
shows a counterexample where the maximizer circuit fails to correctly compute the MPE under the 
empty variable instantiation (all indicators set to \(1\)). Why did \(\adnanmac_3\) not compute the factor MPEs
while \(\adnanmac_2\) did, even though both \(\adnanac_2\) and \(\adnanac_3\)
compute the marginals of their factors? We are almost ready to address this question, but we first need to
introduce another fundamental notion to provide a profound answer.

\subsection*{Complete Subcircuits}

\begin{figure}[tb]
\centering
\includegraphics[width=.25\linewidth]{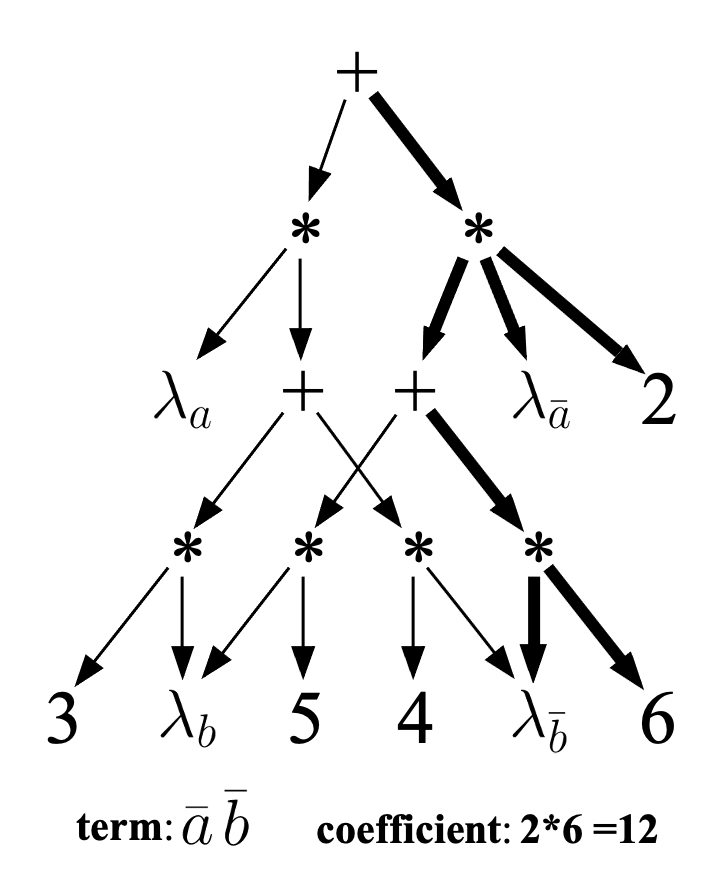}
\quad
\includegraphics[width=.25\linewidth]{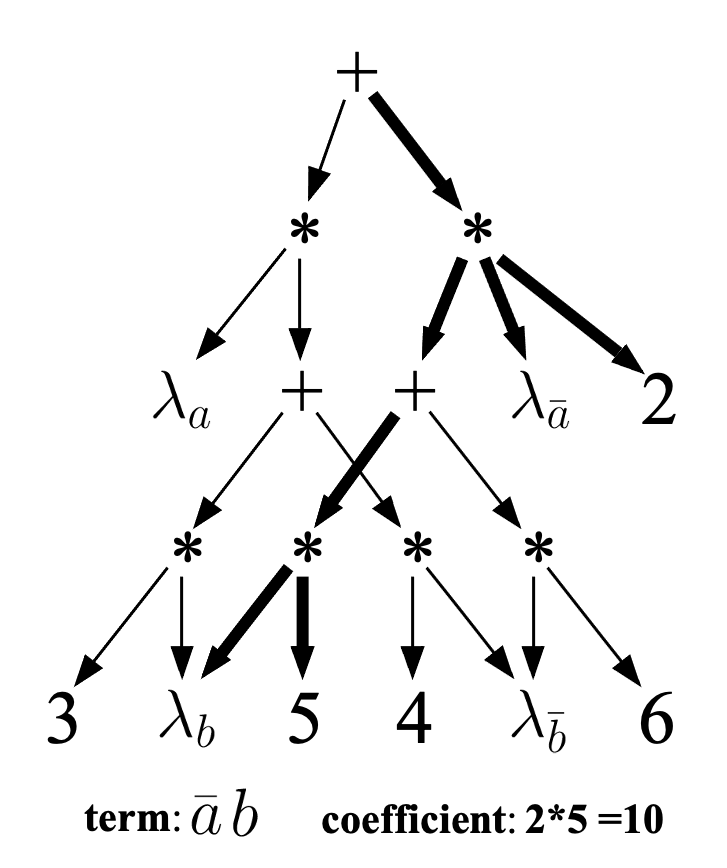}
\caption{Two complete subcircuits (highlighted edges) with their terms and coefficients.
\label{adnan:fig:subcircuits}}
\end{figure}

The fundamental notion of a {\em complete subcircuit} was first introduced and utilized in~\cite{ChanD06}.
We obtain a complete subcircuit by traversing a circuit top-down. When visiting an adder (or
maximizer) node, we choose a single child of the node. When visiting a multiplier node, we choose 
all its children. Figure~\ref{adnan:fig:subcircuits} depicts two examples. A complete subcircuit has a {\em term} and a {\em coefficient.}
The term is the subscripts of indicators appearing in the subcircuit. The coefficient is the product of numbers
appearing in the subcircuit. In Figure~\ref{adnan:fig:subcircuits}, the left subcircuit has term \(\adnann{a},\adnann{b}\) and
coefficient \(12\). The right subcircuit has term \(\adnann{a},b\) and coefficient \(10\).
If a maximizer circuit computes the factor MPEs, we can identify a most likely, complete variable instantiation by constructing
a complete subcircuit as proposed by~\cite{ChanD06}. The procedure is simple
and illustrated on the right of Figure~\ref{adnan:fig:AC2-MPE}. We traverse the circuit top-down.
When visiting a maximizer node, we choose a single child that has the same value as the node. 
When visiting a multiplier node, we choose all its children. The complete subcircuit selected in Figure~\ref{adnan:fig:AC2-MPE}
has term \(\adnann{a},\adnann{b}\) and coefficient \(12\). This means that \(\adnann{a},\adnann{b}\) is a most likely, 
complete variable instantiation and \(12\) is its value.

\subsection*{From Lookup to Reasoning: The Source of Tractability}

Suppose we have an arithmetic circuit that can lookup factor values (i.e., computes the factor).
As discussed earlier, it is generally efficient to construct such circuits (e.g., for the factors specified by probabilistic
graphical models). The fundamental question addressed by~\cite{ChoiDarwiche17} is the following.  
Under what conditions, and {\em why,} will this circuit attain the ability to reason about the factor 
(e.g., compute its marginals or compute its MPEs)? 

The answer rests in the properties satisfied by the arithmetic circuit: decomposability, determinism and smoothness.
We already defined these properties for Boolean circuits. Decomposability and smoothness have identical definitions
on arithmetic circuits when viewing adders/multipliers as or/and gates and indicators \(\lambda_x\) as literals \(\adnaneql(X,x)\). 
Determinism is defined as having at most one non-zero 
input for each adder node, when the circuit is evaluated under any complete variable instantiation.
We already knew from about two decades ago that
if the arithmetic circuit is deterministic, 
decomposable and smooth, it will compute marginals~\cite{DarwicheJACM03}. We also knew later that
such a circuit will compute MPEs~\cite{ChanD06}.
%\footnote{The MPE algorithm we discussed earlier (based on complete subcircuits)
%was proposed in~\cite{ChanD06} but without a proof of correctness. A proof was given later in~\cite{ChoiDarwiche17}
%in the more general setting discussed in this section.}
It was further observed about a decade ago that determinism is not needed for computing marginals, 
leading to a class of arithmetic circuits, known as {\em Sum-Product-Networks (SPNs)}~\cite{PoonD11}, 
which are only decomposable and smooth (\cite{PoonD11} referred to smoothness as {\em completeness}).\footnote{SPNs
placed further structure on the location of circuit parameters, attaching them to the edges/inputs of adder nodes
such that these inputs are multiplied by the corresponding parameters before being added.}
We also came to know recently that relaxing determinism can lead to an exponential reduction in the size of an arithmetic
circuit~\cite{ChoiDarwiche17}.

While determinism is not needed for computing factor marginals, it is needed for the correctness of the linear-time 
MPE algorithm of~\cite{ChanD06} that we discussed earlier.
This was missed in some earlier works~\cite{PoonD11}, which used this algorithm on non-deterministic arithmetic
circuits (i.e., SPNs) without realizing that it is no longer correct. This oversight was noticed in later works~\cite{Peharz16,MauaC17}, which also showed the hardness of computing MPEs without determinism~\cite{Peharz16}.
%\footnote{\cite{Peharz16} proposed a polytime algorithm that converts an SPN 
%into one that is deterministic and smooth (called an \emph{augmented SPN}), but this new
%SPN computes a different factor than the one computed by the original SPN. Hence, its MPEs cannot be generally 
%converted into MPEs of the original SPN.}
The property of determinism was later called {\em selectivity} in the works on SPNs, 
initially in~\cite{peharzlearning}, leading to what has been called {\em Selective SPNs}. 
Since these are arithmetic circuits that satisfy determinism, decomposability and smoothness, they do
compute both marginals and MPEs as was already known earlier~\cite{DarwicheJACM03,ChanD06}.
As we shall see later, determinism plays another important role even when computing MPE is not of interest
as this property allows the compilation of arithmetic circuits from models such as Bayesian networks
without the need to search for circuit parameters (constants).

\begin{figure}[tb]
\centering
\includegraphics[width=.7\linewidth]{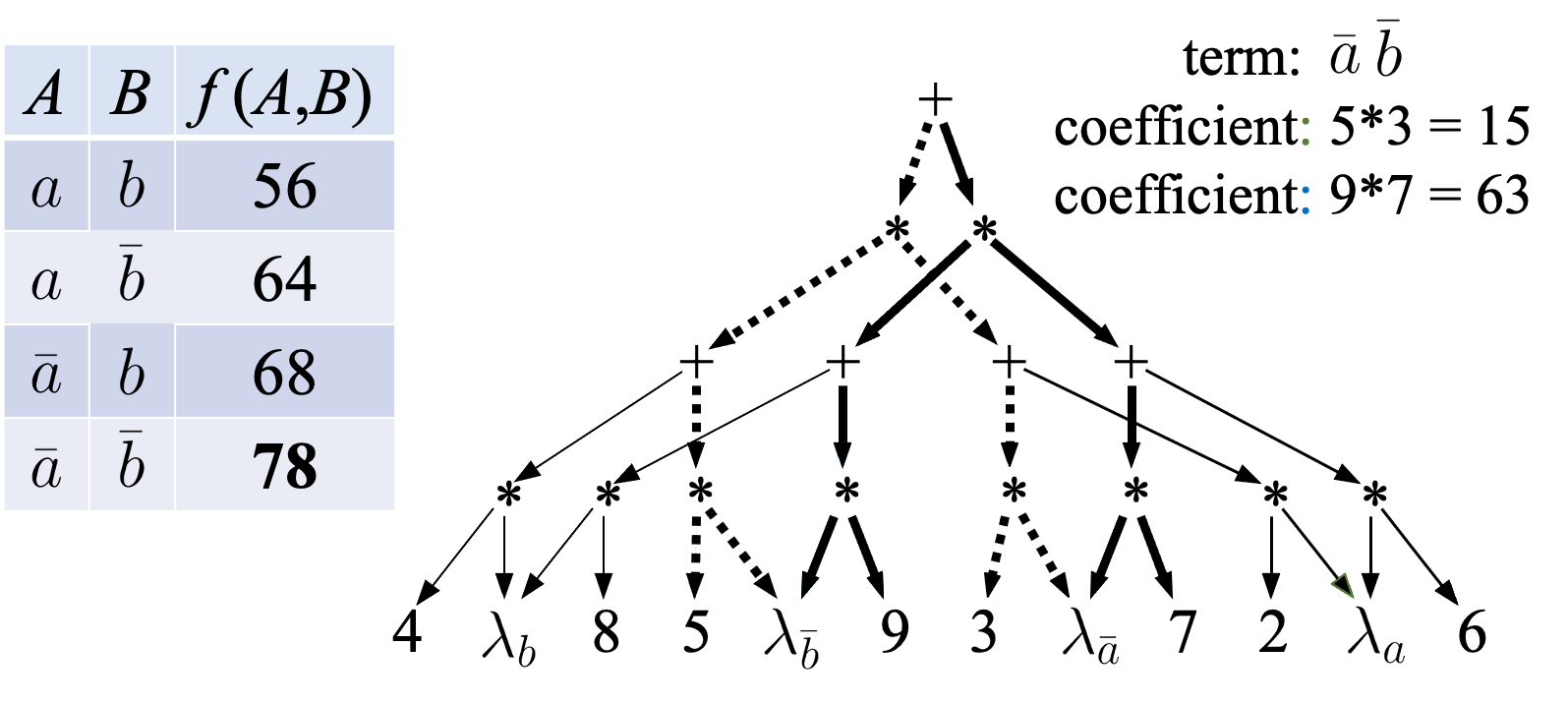}
\caption{A factor and an arithmetic circuit \(\adnanac_3\) that computes it. The circuit is {\bfseries decomposable} 
and {\bfseries smooth} but {\bfseries not deterministic.}
The complete subcircuits with term \(\adnann{a},\adnann{b}\) are highlighted. Their coefficients
are \(15\) and \(63\), which add up to the value \(78\) of instantiation \(\adnann{a},\adnann{b}\).
\label{adnan:fig:AC3-subcircuits}}
\end{figure}

Some of the key insights provided in~\cite{ChoiDarwiche17} related to why the properties of decomposability,
determinism, and smoothness make arithmetic circuits tractable, particularly their ability to peform
reasoning through linear-time circuit traversal. The fundamental notion here is that of a 
complete subcircuit which we discussed earlier. Consider Figure~\ref{adnan:fig:AC3-subcircuits} which depicts 
an arithmetic circuit, \(\adnanac_3\), the factor \(f(A,B)\) it computes, and two complete subcircuits of \(\adnanac_3\).
As mentioned earlier, each complete subcircuit has a term \(\adnanx\) and a coefficient \(c\)
and will be called an \(\adnanx\)-subcircuit. 
Both of the highlighted subcircuits in Figure~\ref{adnan:fig:AC3-subcircuits} 
have \(\adnann{a},\adnann{b}\) as their term and these are the only 
(\(\adnann{a},\adnann{b}\))-subcircuits. 
Their coefficients are \(15\) and \(63\), which add up to \(78\). This is the value assigned by factor \(f\) to the
variable instantiation \(\adnann{a},\adnann{b}\), \(f(\adnann{a},\adnann{b}) = 78\), which is not a coincidence.
As shown in~\cite{ChoiDarwiche17}, decomposability ensures that the term of a complete subcircuit is consistent:
it does not have conflicting variable values. Moreover, smoothness ensures that every variable
is instantiated in the term of a complete subcircuit. Hence, decomposability and smoothness ensure that the
term \(\adnanx\) of a complete subcircuit is a complete variable instantiation. 
Furthermore, for a complete variable instantiation \(\adnanx\), adding up the coefficients of \(\adnanx\)-subcircuits leads
to the value \(f(\adnanx)\) assigned by factor \(f\) to instantiation \(\adnanx\). Finally, when evaluating an arithmetic
circuit at a partial variable instantiation \(\adnane\), the circuit is simply adding up the coefficients of all \(\adnanx\)-subcricuits 
where \(\adnanx\) is compatible with  \(\adnane\), which yields
the correct marginal for that instantiation, \(f(\adnane)\)~\cite{ChoiDarwiche17}. These results provided the first
formal and semantical explanation of why these properties enable an arithmetic circuit
that computes a factor to also reason about that factor, therefore making the circuit tractable.

\begin{figure}[tb]
\centering
\includegraphics[width=\linewidth]{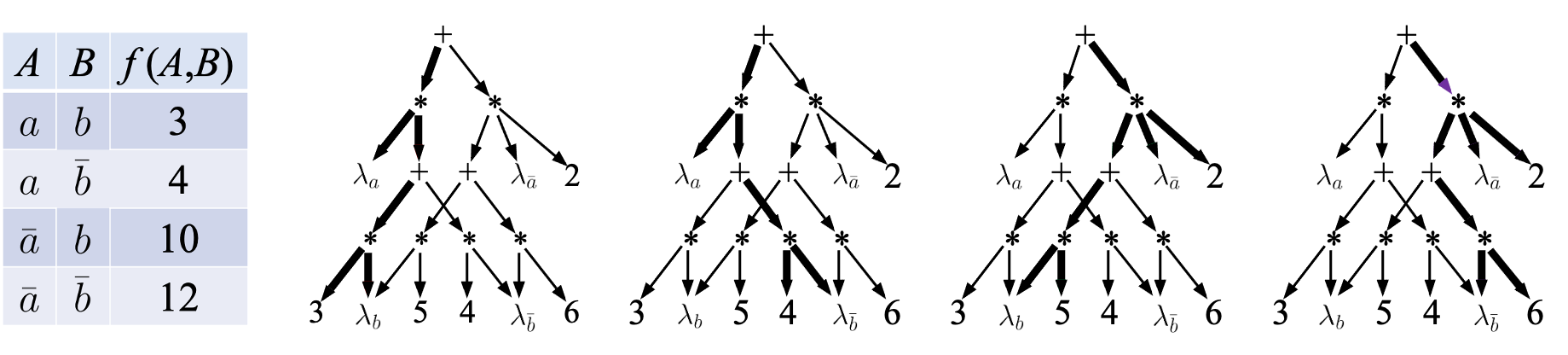}
\caption{A factor and an arithmetic circuit \(\adnanac_2\) that computes it. This circuit is {\bfseries decomposable,} {\bfseries deterministic} and {\bfseries smooth.}
It has four complete subcircuits, in one-to-one correspondence with the factor rows.
\label{adnan:fig:AC2-determinism}}
\end{figure}

We now get to the property of determinism. As shown in~\cite{ChoiDarwiche17},
determinism (with decomposability and smoothness) ensures a one-to-one correspondence 
between complete subcircuits and complete variable instantiations.\footnote{Assuming
no zeros in the circuit or factor; otherwise, the statement of this result is more refined.}
Moreover, the coefficient \(c\) of an \(\adnanx\)-subcircuit corresponds to the value assigned by the factor 
to complete instantiation \(\adnanx\): \(f(\adnanx)=c\). Figure~\ref{adnan:fig:AC2-determinism} depicts an arithmetic
circuit that is deterministic, decomposable and smooth. This circuit has four complete subcircuits, highlighted
in the figure, which are in one-to-one correspondence with the instantiations of variables \(A\) and \(B\).
In the order shown in Figure~\ref{adnan:fig:AC2-determinism}, the subcircuits have terms 
\(a,b;\) \(a,\adnann{b};\) \(\adnann{a},b;\) \(\adnann{a},\adnann{b}\) and coefficients \(3;\) \(4;\) \(10;\) \(12\) which match the rows 
of the factor computed by the circuit.

What determinism does is ensure that each row of the factor (a complete variable instantiation and its value) 
is represented by a single, complete subcircuit. This is essential for the linear-time MPE algorithm of~\cite{ChanD06}
to work properly. Consider Figure~\ref{adnan:fig:AC3-no-MPE} where this algorithm failed to compute the MPE
correctly. The arithmetic circuit in this figure is decomposable and smooth but not deterministic. The MPE
(under no evidence) is the complete instantiation \(\adnann{a},\adnann{b}\) with value \(78\). However, this 
instantiation and its value is not represented by a single, complete subcircuit. Instead, it is represented
by two complete subcircuits with coefficients \(15\) and \(63\) that add up to \(78\) as shown in Figure~\ref{adnan:fig:AC3-subcircuits}.
The MPE algorithm will then return the coefficient \(63\) as shown in Figure~\ref{adnan:fig:AC3-no-MPE}, which is
not correct.

\begin{figure}[tb]
\centering
\includegraphics[width=\linewidth]{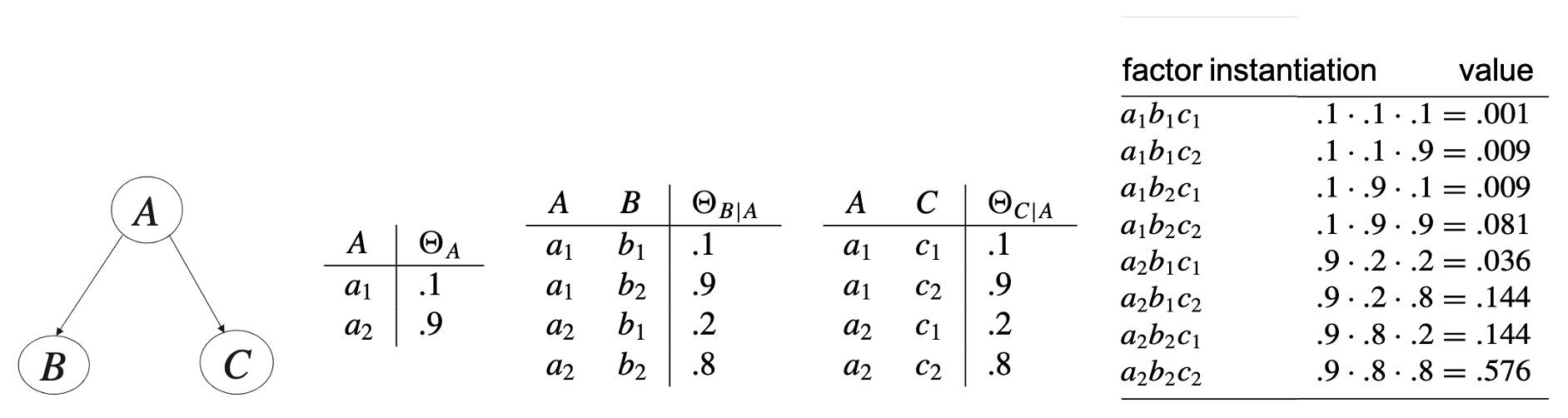}
\caption{A Bayesian network structure, its conditional probability tables (factors \(\Theta_A\), \(\Theta_{B|A}\), \(\Theta_{C|A}\)) 
and the network distribution (factor \(f\)).
The distribution is a product of conditional probability tables, 
\(f = \Theta_A \Theta_{B|A}\Theta_{C|A}\).
\label{adnan:fig:bn-factor}}
\end{figure}

\subsection*{More on the Need for Determinism}

While determinism is not needed for computing marginals, it can be critical for compiling arithmetic circuits 
from models as we illustrate next.
An arithmetic circuit has both a structure and parameters (the constants appearing as circuit inputs).
When compiling an arithmetic circuit from a model such as a Bayesian network, one is guaranteed to
find a circuit that computes marginals and whose parameters are restricted to the network parameters, 
if the circuit is deterministic in addition to being decomposable and smooth. 
However, this guarantee is lost if the circuit is only decomposable
and smooth (an SPN) so one must also search for circuit parameters in this case~\cite{ChoiDarwiche17}. 
To elaborate further on this result and its significance, consider  
the Bayesian network in Figure~\ref{adnan:fig:bn-factor} which has four
distinct parameters \(\Theta = \{.1,.2,.8,.9\}\). Consider now the distribution (factor \(f\)) specified by this network.
Each value in this factor can be expressed as a product of some parameters that come from \(\Theta\) (this
is guaranteed by the semantics of Bayesian networks and probabilistic graphical models more generally).
We say in this case that the set of parameters \(\Theta\) is {\em complete} for the factor \(f\) (i.e., we can express
the values of \(f\) using products of numbers from \(\Theta\)).
If a set of parameters \(\Theta\) is complete for a factor \(f\), 
there is always an arithmetic circuit that computes factor \(f\), that is deterministic, decomposable 
and smooth, and whose parameters come from the set \(\Theta\).
Therefore, when constructing (compiling) such a circuit, one does not need to search for circuit parameters
beyond the set \(\Theta\). This no longer holds without determinism so the compilation process must
also search for circuit parameters in addition to its structure---we do not know of any such compilation algorithm 
today.\footnote{This result was shown in Theorem~6 of~\cite{ChoiDarwiche17} for circuits that compute Boolean 
factors \(f(\adnanX)\). 
The parameters \(\Theta = \{0,1\}\) are complete for these factors since each row has a value in \(\{0,1\}\).
If the circuit is decomposable and smooth but not deterministic, then some row \(\adnanx\) with \(f(\adnanx)=1\)
must be split over at least two \(\adnanx\)-subcircuits with non-zero coefficients. Since the coefficients of these 
\(\adnanx\)-subcircuits must add up to \(1\), there must exist an \(\adnanx\)-subcircuit whose coefficient is in the
open interval \((0,1)\) so the circuit must have a parameter not in \(\{0,1\}\).}
Consider the Bayesian network in Figure~\ref{adnan:fig:bn-factor} and the following arithmetic circuit
which does not require searching for parameters:
\[ 
{.001} * \lambda_{a_1}*\lambda_{b_1}*\lambda_{c_1} + 
{.009} * \lambda_{a_1}*\lambda_{b_1}*\lambda_{c_2} + \ldots +
{.576} * \lambda_{a_2}*\lambda_{b_2}*\lambda_{c_2}.
\]
This has been called the {\em network polynomial} in~\cite{DarwicheJACM03} 
and the {\em factor polynomial} in~\cite{ChoiDarwiche17}. 
Each monomial in this polynomial 
corresponds to a factor row: its indicators capture the row's instantiation and its coefficient
captures the row's value (which is a product of Bayesian network parameters). 
For example, the last monomial \({.576} * \lambda_{a_2}*\lambda_{b_2}*\lambda_{c_2}\)
captures instantiation \(a_2,b_2,c_2\) and value \(.576 = .9*.8*.8\).
Viewed as a depth-two arithmetic circuit, this polynomial is deterministic, decomposable and smooth, 
and all its paramaters come from the Bayesian network parameters.
It is also exponentially sized so it is mostly of theoretical interest such as illustrating the
result we just discussed.\footnote{For another theoretical result with substantial practical implications,
we note that the polynomial
partial derivates with respect to indicators and parameters correspond to 
useful probabilistic quantities which can be computed by backpropagation on any equivalent arithmetic 
circuit. These quantities correspond to marginals of additional variable instantiations. They allow one
to avoid further evaluations of the circuit which can lead to significantly
more efficient reasoning; see~\cite{DarwicheJACM03} for details.}
Existing compilation algorithms construct more compact arithmetic circuits without searching for
parameters, assuming the model parameters are known. When the model parameters are not known, 
these algorithms compile arithmetic circuits with symbolic parameters that can be easily replaced with
numeric parameters that may be learned from data. This is a topic 
we shall discuss in Section~\ref{adnan:sec:compile}.

It is worth noting that the original treatment of arithmetic circuits in~\cite{DarwicheJACM03} started where our 
current treatment has ended in the previous paragraph. That is, the starting point of~\cite{DarwicheJACM03} was the 
notion of a network polynomial which fully captures a distribution. The notion of an arithmetic circuit was then 
introduced as a tool that can be used to compactly represent the exponentially-sized network polynomial. 
This formulation was facilitated by the fact that~\cite{DarwicheJACM03} also started by assuming the existence 
of a model (a Bayesian network), which fully defines the network polynomial. 
This can no longer be assumed when handcrafting circuits or learning them from data, which prompted the modern 
treatment in~\cite{ChoiDarwiche17}. We finally note that arithmetic circuits which compute distributions and
their marginals can be used to reason about uncertain evidence, also called {\em soft evidence,} in addition
to evidence about continuous variables with local, univariate densities. This can be done by setting the 
circuit indicators to appropriate real values as discussed in~\cite{ai/ChanD05} and~\cite[Section 3.7]{Darwiche09}, 
respectively.\footnote{\label{adnanfoot:soft-evidence}Consider a variable \(X\) with values \(x_1, \ldots, x_n\). 
Uncertain evidence on variable \(X\) can be modeled 
as certain (hard) evidence on some noisy sensor \(\eta\) that is connected to variable \(X\), where the noise is 
specified by the likelihoods of \(x_i\) given \(\eta\), \(\adnanpr(\eta | x_i)\). 
We can assert this uncertain evidence on variable \(X\) by setting each indicator \(\lambda_{x_i}\)
to its corresponding likelihood \(\adnanpr(\eta | x_i)\). Let \(a\) be the circuit value under this indicator setting and let \(b\) be
its value when we further set indicators according to additional evidence \(\adnane\). 
Then \(\adnanpr(\adnane | \eta) = b/a\). Moreover, this conditional probability is invariant to the specific
values of indicators as long as their relative ratios match the corresponding ratios of 
likelihoods; see Theorem~2 in~\cite{ai/ChanD05} and~\cite[Section 3.6]{Darwiche09} for details.
This technique is known as the method of {\em virtual evidence} in the context of Bayesian  networks~\cite{Pearl88b}.
The treatment of continuous variables is similar except that the values of indicators are determined by
the observed value of the continuous variable and its density functions. 
This is discussed at length in~\cite[Section 3.7]{Darwiche09}.}

\subsection*{Circuits that Reason about Constrained Factors}

\begin{figure}[tb]
\centering
\includegraphics[width=.4\linewidth]{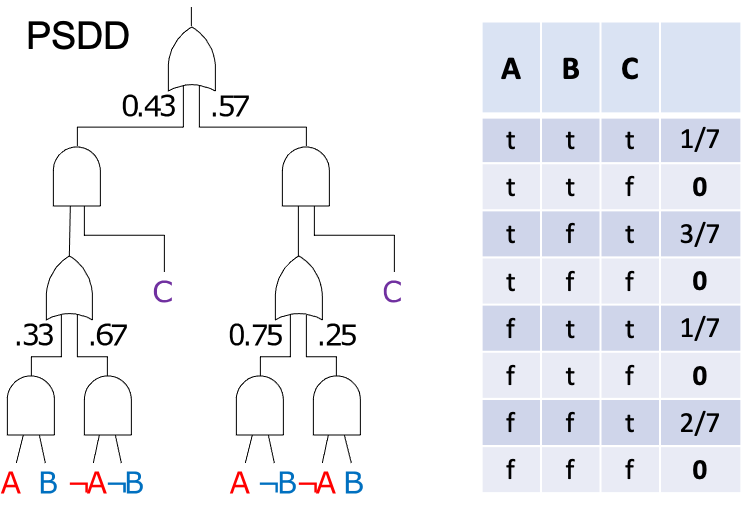}
\qquad
\includegraphics[width=.2\linewidth]{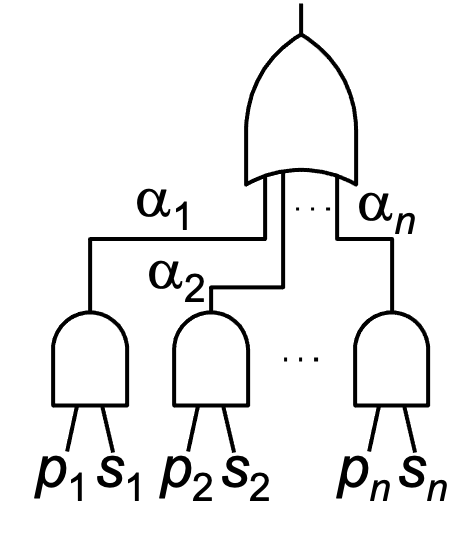}
\includegraphics[width=.33\linewidth]{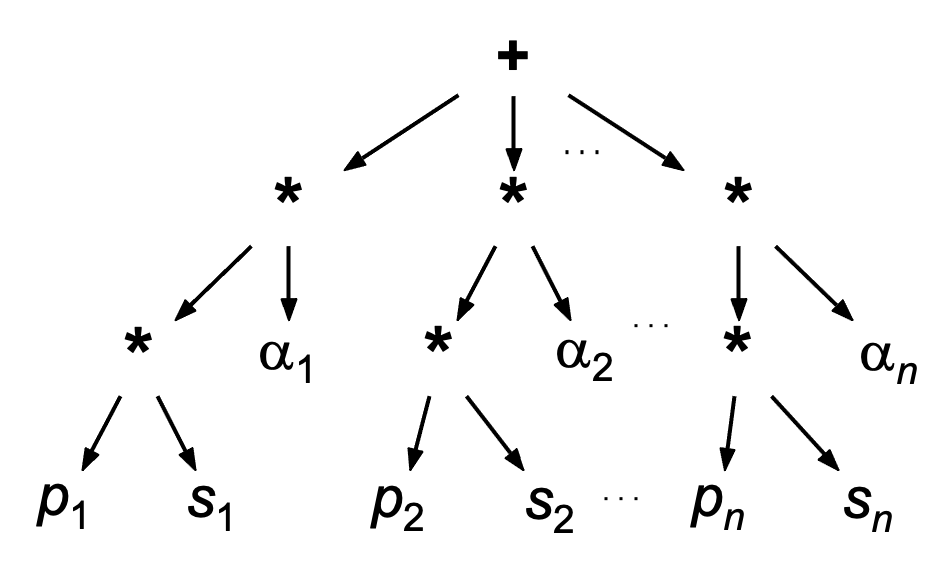}
\caption{Left: A PSDD and the constrained factor it computes. 
The factor computed by this PSDD will always assign zero values to rows in which \(\adnaneql(C,f)\),
regardless of how the PSDD is parameterized.
Right: A PSDD fragment and its corresponding arithmetic circuit fragment.
Boolean variables and their negations in the PSDD (e.g., \(A\) and \(\neg A\)) are converted to indicators
in the arithmetic circuit (\(\lambda_{A}\) and \(\lambda_{\neg A}\)).
\label{adnan:fig:psdd-factor}}
\end{figure}

We now turn to a fundamental class of arithmetic circuits, known as {\em PSDDs,}\footnote{PSDD
stands for {\em Probabilistic Sentential Decision Diagram.} A PSDD represents a probability distribution
but can also be used to represent and reason about general factors~\cite{ShenCD16}.} 
which can reason about {\em constrained factors}~\cite{KisaBCD14}. These are factors in which 
some rows have fixed zero values so they define mappings from a {\em subset} of variable instantiations 
into non-negative numbers; see Figure~\ref{adnan:fig:psdd-factor}. 
Constrained factors, particularly ones representing distributions, have many applications. 
For example, when learning arithmetic circuits that represent distributions from data, one may have  
domain knowledge that rules out certain states of the world so we need to ensure that any learned circuit will assign
a zero probability to such infeasible states. One may also want to induce or
learn distributions over combinatorial (or structured) objects such as total and partial rankings~\cite{ChoiVdbDarwiche15},
graphs~\cite{SDDgraphs17}, game plays~\cite{ChoiTavabiDarwiche16}, routes on a 
map~\cite{aips/LingCK21,ChoiTavabiDarwiche16,ChoiSD17,ShenChoiDarwiche17,aaai/ShenGDC19} and subsets of 
objects~\cite{ShenChoiDarwiche17}. Distributions over these kind of objects have been captured
using constrained factors and reasoned about using PSDDs as done in the works mentioned in the previous sentence. 

\begin{figure}[tb]
\centering
\includegraphics[width=.6\linewidth]{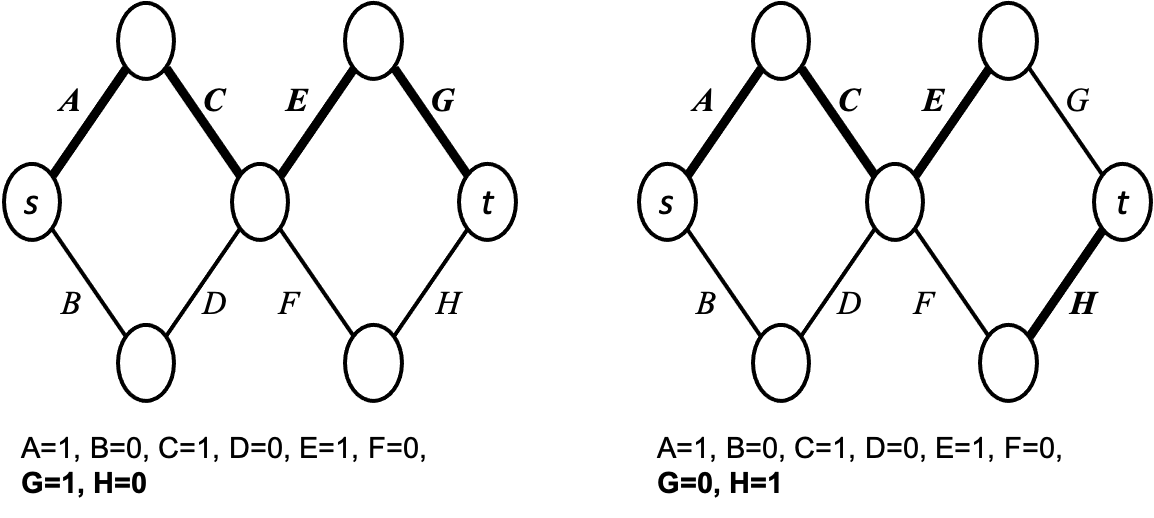}
\caption{Modeling routes between source \(s\) and destination \(t\) as variable
instantiations that satisfy some conditions. Each edge between two locations is
modeled using a Boolean variable which is set to \(1\) if the edge is on a route
and set to \(0\) otherwise. The variable instantiation on the left corresponds to
a valid route. The variable instantiation on the right corresponds to an invalid (disconnected) route.
\label{adnan:fig:routes}}
\end{figure}

For an example of how constrained factors can be used to model combinatorial and structured objects, 
consider Figure~\ref{adnan:fig:routes} where the goal is to define a distribution over routes from source~\(s\) 
to destination~\(t\). Each edge on
the map is modeled by a binary variable, \(A, \ldots, H.\) A route can be modeled using a variable
instantiation that sets the variables of its edges to \(1\) and all other variables to \(0\).
Clearly, some variable instantiations do not correspond to valid routes so these get assigned 
probability \(0\); see the right of Figure~\ref{adnan:fig:routes}. The corresponding constrained factor is then 
guaranteed to induce a distribution over only valid routes from source \(s\) to destination \(t\).
Similar techniques can be used to represent other combinatorial or structured objects as long as one can
define the conditions that specify variable assignments which correspond to the  
objects of interest.

A constrained factor is defined by specifying feasible variable instantiations and their values.
The main insight behind PSDDs is to specify feasible instantiations using an SDD, 
which is a tractable Boolean circuit that we discussed in Section~\ref{adnan:sec:BC}; see 
Figure~\ref{adnan:fig:psdd-factor} (left). The SDD should evaluate feasible instantiations to \(1\)
and infeasible instantiations to \(0\) and is normally obtained through a compilation process
of domain constraints. A distribution over the
feasible instantiations can then be defined by assigning (local) distributions to the inputs of or-gates in the SDD,
leading to a PSDD as shown in Figure~\ref{adnan:fig:psdd-factor}. 

\begin{figure}[tb]
\centering
\includegraphics[width=\linewidth]{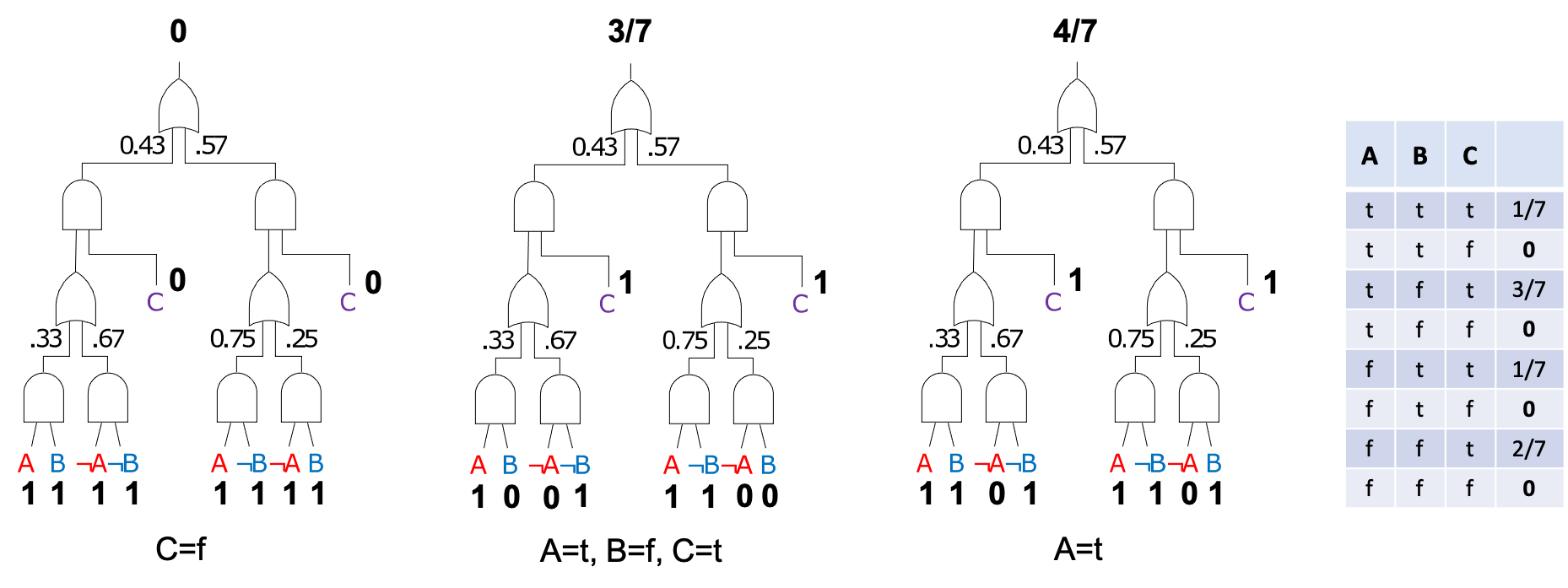}
\caption{Evaluating a PSDD at different variable instantiations (complete and partial). 
The PSDD will evaluate to \(0\) at any variable instantiations that does not satisfy the underlying SDD circuit.
This is guaranteed regardless of how the PSDD is parameterized.
\label{adnan:fig:psdd-eval}}
\end{figure}

The PSDD is a highly structured arithmetic circuit that comes with strong guarantees.
We can convert a PSDD into an arithmetic circuit as shown in Figure~\ref{adnan:fig:psdd-factor} (right)
by removing reference to the underlying SDD circuit. The resulting arithmetic circuit is guaranteed
to not only satisfy determinism, decomposability and smoothness but to also satisfy the stronger
properties of structured decomposability and partitioned determinism which are forced by the
underlying SDD circuit which satisfies these properties. As a result, the induced arithmetic circuit
enjoys many desirable properties. First, the circuit is guaranteed to produce a zero value for 
any infeasible variable instantiation regardless of how we set the PSDD parameters. 
Figure~\ref{adnan:fig:psdd-eval} depicts example evaluations of a PSDD circuit that can be used
to gain insights into this guarantee. Moreover, the PSDD can compute marginals and MPEs
by virtue of being deterministic, decomposable and smooth~\cite{KisaBCD14}. 
With an appropriate vtree for the SDD circuit, it can also compute expectations~\cite{kr/OztokCD16}. 
One can also learn the maximum-likelihood 
parameters of a PSDD in closed form when the data is complete~\cite{KisaBCD14}. An EM algorithm has also 
been developed for learning PSDD parameters from incomplete data, including highly structured data
in which examples can be specified using Boolean constraints, not just variable instantiations~\cite{ChoiVdbDarwiche15}.
PSDDs also play a key role in probabilistic reasoning as shown in~\cite{ShenCD16}.

\begin{figure}[tb]
\centering
\includegraphics[width=.75\linewidth]{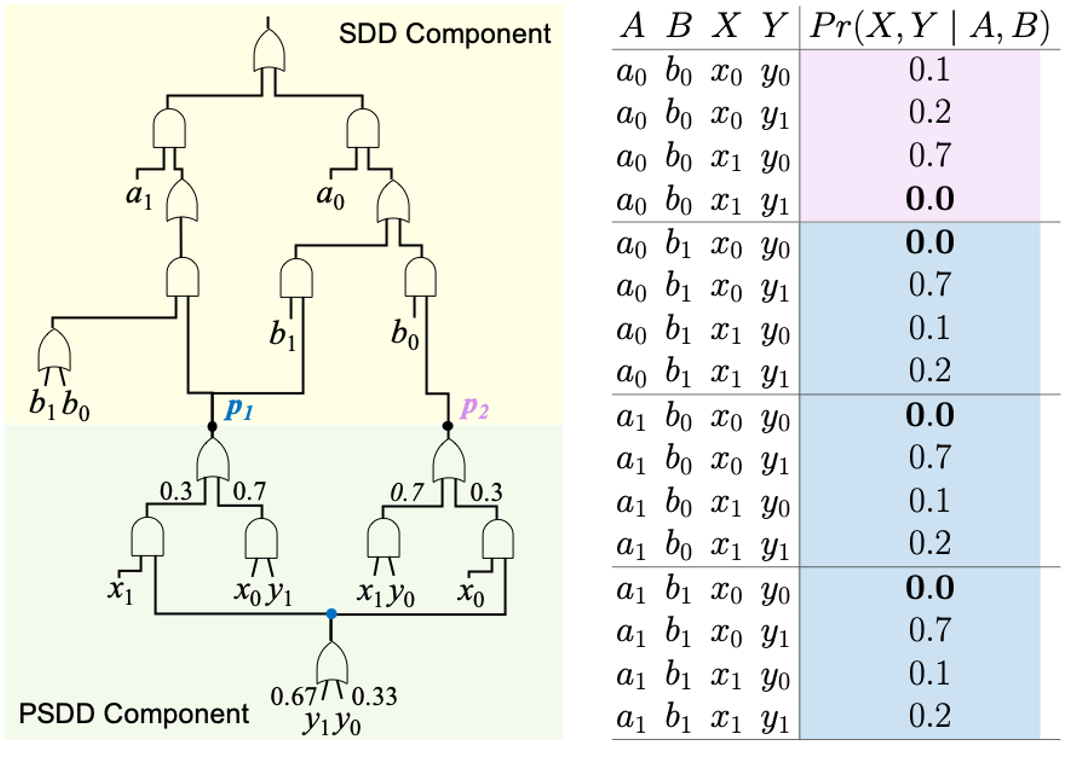}
\caption{A conditional PSDD composed of an SDD circuit and two PSDDs
that share structure. The SDD defines two constrained factors over variables \(X\) and
\(Y\) depending on the state of variables \(A\) and \(B\). These constrained
factors are specified by PSDDs that share structure.
\label{adnan:fig:c-psdd}}
\end{figure}

Another key development has been the introduction of {\em conditional PSDDs,} which can
compute and reason about factors that are {\em conditionally constrained}~\cite{ShenCD18}.
These are factors in which variables are partitioned into two sets, \(\adnanX\) and \(\adnanY\), where
the infeasible instantiations of \(\adnanY\)-variables are conditioned on the state of \(\adnanX\)-variables.
Conditional PSDDs allow one to integrate {\em conditional constraints} into tractable arithmetic
circuits. For example, which routes are feasible in a neighborhood may be conditional on
how we enter and exit that neighborhood~\cite{aaai/ShenGDC19}. Figure~\ref{adnan:fig:c-psdd} depicts a
conditional PSDD, which is composed of an SDD circuit and a set of PSDDs that share structure.

\subsection*{Further Extensions}

The theory of tractable arithmetic circuits has been developed and explored in various additional ways. For example, 
additional classes of arithmetic circuits have been studied using further combinations of tractable circuit properties, 
such as the combination of structured decomposability and smoothness~\cite{pgm/DangVB20}. 
Other reasoning tasks have also been considered
which are enabled by the different properties of tractable circuits; see, e.g.,~\cite{OpAtlas21}. Some of the new
extensions and works have referred to arithmetic circuits as {\em probabilistic circuits,} while reserving
the term ``arithmetic circuits'' to deterministic, decomposable and smooth circuits.
This is in contrast to earlier treatments such as~\cite{ShenCD16,ChoiDarwiche17} which did not
tie this term to specific properties or exclusively to probability distributions. A fundamental new
extension has been the class of {\em Testing Arithmetic Circuits (TACs)} which select their parameters dynamically,
based on circuit inputs through the inclusion of {\em tests} in the arithmetic circuit~\cite{pgm/ChoiD18,ijar/ChoiWD19}. 
Tractable arithmetic circuits represent multilinear functions (of indicators)~\cite{DarwicheJACM03}.
Testing arithmetic circuits represent {\em piecewise} multilinear functions so they are universal function
approximators like neural networks~\cite{ijar/ChoiWD19}. This new class of arithmetic circuits can be used to recover from
some modeling errors that lead to compiled circuits that may not be expressive
enough to fit the data generated by a mechanism that has been modeled incorrectly~\cite{icml/ShenHCD19}.
Recent theoretical results have shown that full recovery from some modeling errors 
is possible under certain conditions~\cite{icml/HuangD21}.
We finally note that some of the key properties of tractable circuits, including decomposability and determinism,
have been exploited for tractable reasoning and learning in a more general, semiring setting; see, e.g.,~\cite{Kimmig12,KimmigJAL16,FriesenD16,ijar/BelleR20}.

\section{Compiling Models into Tractable Circuits}
\label{adnan:sec:compile}

We will next provide an overview of methods for constructing tractable circuits through a process of compilation.
We will provide a brief summary of methods for tractable Boolean circuits and elaborate 
more on methods for tractable arithmetic circuits.

Tractable Boolean circuits are normally compiled from Boolean formulas that represent logical knowledge or constraints. This is done
using systems known as {\em knowledge compilers,} such as
\adnandf~\cite{DBLP:conf/ijcai/LagniezM17},
\adnanctd~\cite{Darwiche04},
\adnancudd,
\adnanmctd~\cite{OztokD14b,OztokD18}, 
\adnandsharp~\cite{DBLP:conf/ai/MuiseMBH12} and
the \adnansdd\ library~\cite{ChoiDarwiche13}.\footnote{\adnandf: \url{http://www.cril.univ-artois.fr/kc/d4}; 
\adnanctd: \url{http://reasoning.cs.ucla.edu/c2d};
\adnancudd: \url{https://davidkebo.com/cudd};
\adnanmctd: \url{http://reasoning.cs.ucla.edu/minic2d};
\adnandsharp: \url{https://bitbucket.org/haz/dsharp};
\adnansdd\ library: \url{http://reasoning.cs.ucla.edu/sdd}.
A Python wrapper of the \adnansdd\ library is available at \url{https://github.com/wannesm/PySDD}
and an NNF-to-SDD circuit compiler, based on \adnanpysdd, is available at \url{https://github.com/art-ai/nnf2sdd}.}
Knowledge compilers can be categorized as {\em top-down} or {\em bottom-up.} Top-down compilers 
are based on keeping a {\em trace} of the exhaustive DPLL algorithm~\cite{HuangD07}
and they generally incorporate advanced SAT techniques~\cite{sat/SangBBKP04,OztokD18}.
These compilers normally operate on Boolean formulas in conjunctive normal form, tend to have better space 
complexity, and include \adnandf, \adnanctd\ and \adnandsharp\ which yield Decision-DNNFs, and \adnanmctd\ which yields SDDs.
Bottom-up compilers incrementally compile a formula, by first compiling its components and then combining these 
compilations. They tend to operate on more general classes of Boolean formulas, demand
more space, and include \adnancudd\ and 
the \adnansdd\ library, which yield OBDDs and SDDs, respectively.

Turning next to the compilation of tractable arithmetic circuits, we first note that  
PSDDs, which represent distributions, are somewhat special as these arithmetic circuits are based on SDDs, which are 
Boolean circuits. Hence, the structure of a PSDD is obtained by first compiling a Boolean formula
into an SDD, which is then
parameterized to yield a PSDD. The PSDD parameters are typically obtained through a learning process based on
complete or incomplete data~\cite{KisaBCD14,ChoiVdbDarwiche15}. 
The Boolean formula that triggers this compilation process usually defines
feasible states of the world, or characterizes structured
(combinatorial) objects such as routes, graphs, and rankings.\footnote{See 
\url{https://github.com/art-ai/pypsdd};
\url{http://reasoning.cs.ucla.edu/psdd}; and
\url{https://github.com/hahaXD/hierarchical_map_compiler} for some related tools.}

Arithmetic circuits that satisfy determinism, decomposability
and smoothness, also known as ACs, are typically compiled from probabilistic models such as Bayesian
networks and probabilistic logic programs. We will only discuss the former
but see~\cite{bnaic/ManhaeveDKDR19,tplp/FierensBRSGTJR15,pkdd/DriesKMRBVR15} 
for some examples of the latter. Recall again that the structure of arithmetic circuits that satisfy only decomposability and
smoothness (i.e., SPNs) is normally handcrafted or learned from data since 
relaxing determinism complicates the compilation of these structures from models as discussed in the previous section.

\begin{figure}[tb]
\centering
\includegraphics[width=.52\linewidth]{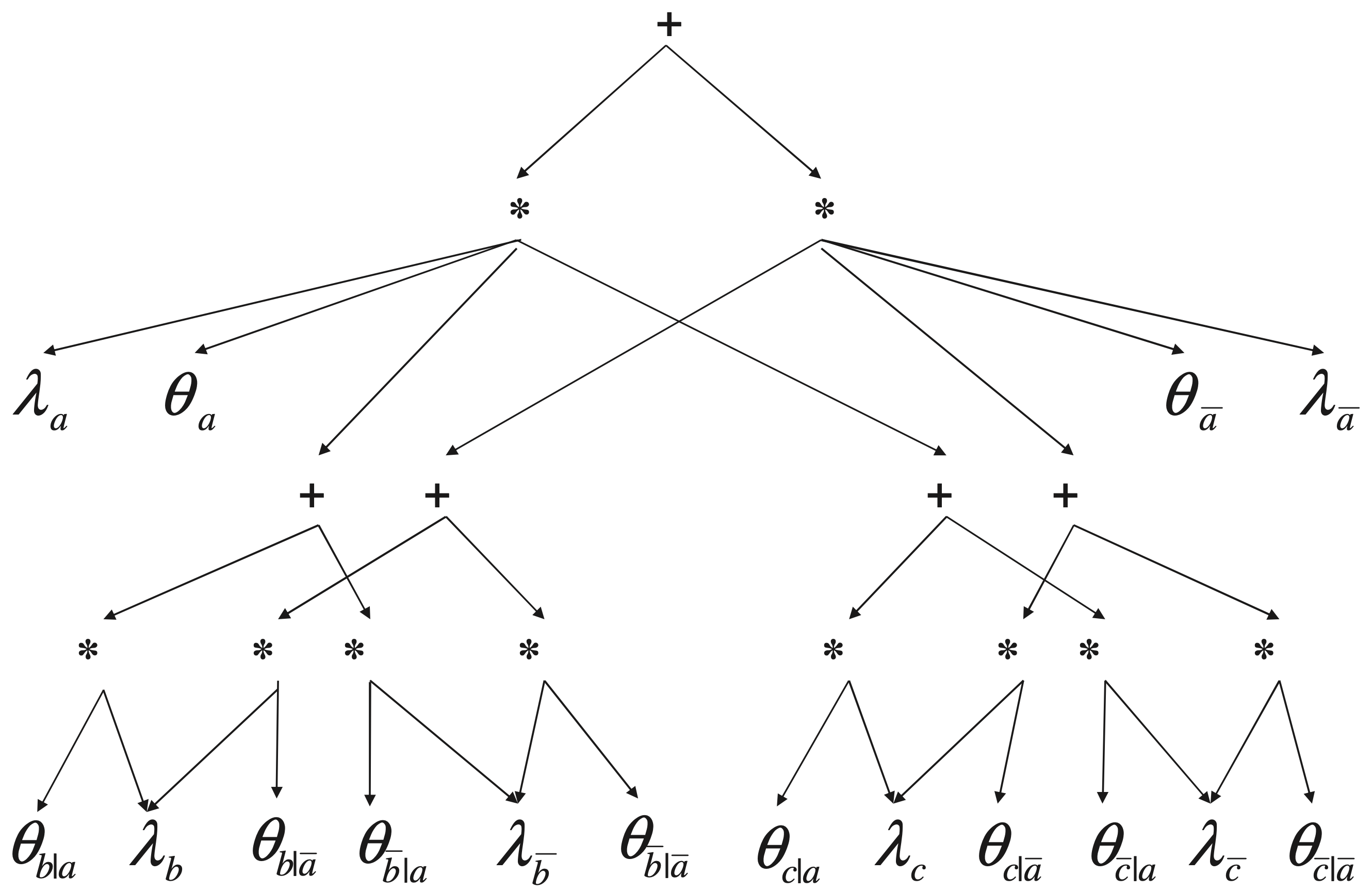}
\quad
%\includegraphics[width=.2\linewidth]{figures/ABC-BN}
%\quad
\includegraphics[width=.44\linewidth]{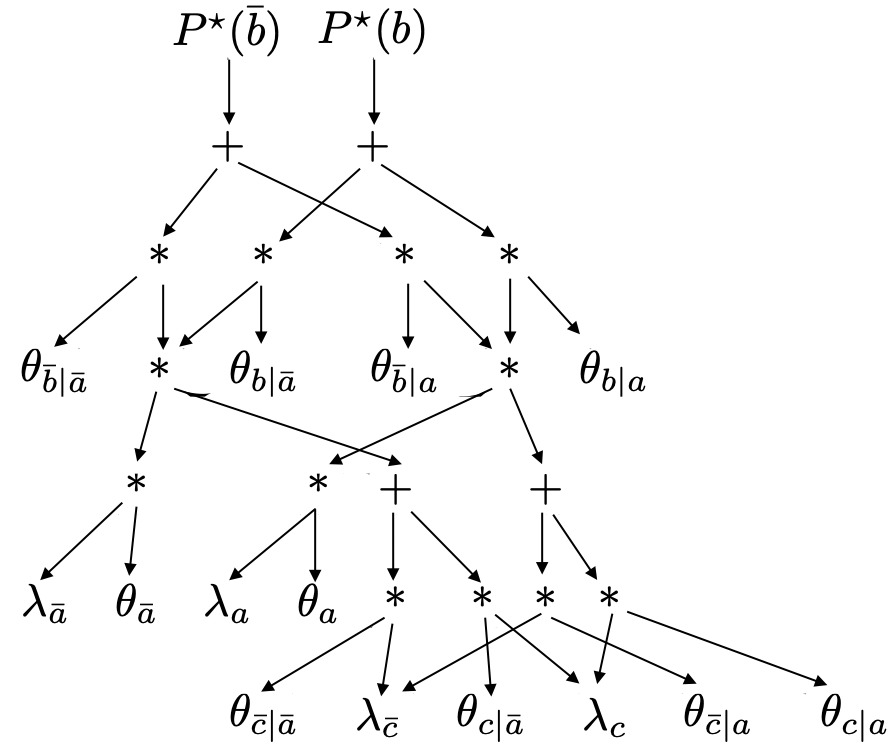}
\caption{Compiling the Bayesian network \(B \leftarrow A \rightarrow C\) into tractable arithmetic circuits.
The circuit on the left computes the probability of (partial) evidence on variables \(A\), \(B\) and \(C\).
The circuit on the right computes the distribution of variable \(B\) given (partial) evidence on 
variables \(A\) and \(C\) (after normalization); this circuit targets a particular query, \(\adnanpr(B|.)\).
In both cases, the circuit parameters correspond to Bayesian network parameters so they can be
fetched from the Bayesian network, if known; otherwise, they can be learned from data.
\label{adnan:fig:BN2ACs}}
\end{figure}

The compilation of Bayesian networks (and probabilistic graphical models more generally)
is done either directly, or indirectly as shown in Figure~\ref{adnan:fig:pipeline}.
{\em Indirect methods} are based on reductions to weighted model counting~\cite{ai/ChaviraD08}. 
The model is first {\em encoded} into a Boolean formula with weights on literals, 
as initially proposed in~\cite{kr/Darwiche02,DarwicheJACM03}.
The Boolean formula is then compiled into a smooth d-DNNF circuit (a tractable Boolean
circuit), from which a tractable arithmetic circuit is extracted. The size of the final arithmetic circuit 
depends on the size of compiled Boolean circuit, which is determined by the quality of used encoding scheme and knowledge compiler. 
These encoding schemes capture both the {\em global structure} of the model (its topology) and its {\em local structure} (parameters).
Their efficacy relies particularly on how well they encode local structure;
that is, the extent to which they capture the properties of model parameters. The first encoding scheme~\cite{kr/Darwiche02}
was somewhat basic in nature, but it was later followed by more refined and potent encodings 
in~\cite{aaai/SangBK05,ijcai/ChaviraD05,sat/ChaviraD06}. A comparative discussion of various
encoding schemes, including some more recent ones, is given in~\cite{uai/DilkasB21}.
The above approaches have traditionally used top-down compilers of Boolean formulas into tractable Boolean circuits; for example,
the \adnanace\ system~\cite{ai/ChaviraD08} used the top-down \adnanctd\ knowledge compiler.
More recently, bottom-up compilation approaches have also been proposed, which are
based on compiling the factors of a probabilistic graphical model into tractable
circuits and then combining these compiled circuits in a bottom-up fashion~\cite{ecsqaru/ChoiKD13,ShenCD16}.

{\em Direct methods} for compiling models into tractable arithmetic circuits are simpler but they tend to be less
effective, with some exceptions. These methods include the extraction of a tractable arithmetic circuit from 
the structure of a jointree for the model~\cite{DarwicheJACM03}. They also include methods based on the variable 
elimination algorithm~\cite{Chavira.Darwiche.Ijcai.2007,DarwicheECAI20b}; see also~\cite[Chapter 12]{Darwiche09}.
The most recent of these methods~\cite{DarwicheECAI20b} is based on variable elimination
and stands out for a number of reasons. First, the method compiles circuits that target a specific class of queries,
specified by evidence variables (input) and a query variable (output), which is meant to 
facilities supervised learning from labelled data; see Figure~\ref{adnan:fig:BN2ACs}.
Second, this recent method compiles arithmetic circuits in the form of computation graphs in which nodes represent
tensor operations instead of arithmetic operations, allowing parallelization during parameter learning and 
inference.\footnote{See also~\cite{icml/PeharzLVS00BKG20,molina2019spflow} for related representations 
of circuits with handcrafted or learned structures.} 
Third, this method computationally exploites a new type of local structure: {\em functional dependencies}
whose identities are {\em unknown.}  
We have a functional dependency between a node and its parents in the Bayesian network
when the state of the node is a function of the states of its parents (i.e., there is no uncertainty). 
A functional dependency is unknown when we do not know the identity of this function. This is significant in a learning context where the 
goal is to learn such functions from data. This is also significant for causal inference where functional dependencies are known
as {\em causal mechanisms} which are typically unknown~\cite{causalityAC}.
This latest algorithm~\cite{DarwicheECAI20b}
was motivated by the use of tractable arithmetic circuits for supervised learning, as in neural
networks, but where the structure of the circuit is compiled from a model instead of
being handcrafted as is normally the case with neural networks. Recent results~\cite{pgm/ChenCD20} 
have shown the promise of this approach to supervised learning as it provides a principled method 
for embedding background knowledge into the learning process (independence 
constraints, logical constraints, known parameters, and unknown functional dependencies).

We close this section by the following remark on the interplay between the compilation of circuit structure
and the learning of circuit parameters. 
Initially, the goal of compilation algorithms was to facilitate reasoning since various forms of inference can be performed in
linear time on the compiled circuits. In this context, the assumption was that one already knows the model parameters before
the compilation process starts.
Later, however, these compilation algorithms were used to compile only the circuit structure and then
coupled with additional algorithms that learn the circuit parameters from data. 
The key observation which permits this expanded role is that
the compilation process can be conducted even if we do not know the model parameters. In particular,
the substitution of model parameters into the compiled circuit can be postponed to the extraction 
phase of Figure~\ref{adnan:fig:pipeline}. 
Hence, if a parameter is unknown, it can be kept in symbolic form during the extraction phase and 
then learned from labeled or unlabeled data after the compilation process is concluded; e.g., 
as in~\cite{DarwicheECAI20b,pgm/ChenCD20} and~\cite{KisaBCD14,ChoiVdbDarwiche15}, respectively.
This is possible since determinism (with decomposability and smoothness) allows one to
compile circuits with parameters that correspond to model parameters as discussed in the
previous section; see also the circuits in Figure~\ref{adnan:fig:BN2ACs} and~\cite{causalityAC} for a recent
exposition of how this is done. Interestingly enough, determinism also facilitates
the learning of circuit parameters, allowing one to learn parameters in closed form under complete data, in addition
to allowing a linear-time evaluation of the EM update equations
under incomplete data.\footnote{For complete data, this follows directly from the same
result for Bayesian networks. For incomplete data, a second pass (backpropagation) on the circuit provides marginals
over families (variables and their parents)~\cite{DarwicheJACM03} which is all that one needs to evaluate the EM update equations~\cite[Eq~17.7]{Darwiche09}.}
This further highlights the importance of determinism, which also holds for tractable arithmetic circuits with
handcrafted or learned structures.\footnote{For arithmetic circuits with handcrafted or learned structures, 
determinism can also facilitate parameter learning if these parameters are properly located as in~\cite{peharzlearning}
and~\cite{KisaBCD14}. For further perspective, see~\cite{nips/ZhaoPG16} for a treatment of parameter learning when the arithmetic 
circuit does not satisfy determinism.}

\section{Concluding Remarks}
\label{adnan:sec:conclusion}

Tractable Boolean and arithmetic circuits have been evolving into a computational and semantical backbone for modern approaches 
that aim to combine knowledge, reasoning, and learning. The different modes of constructing these circuits---through
compilation, handcrafting and learning---have further contributed to their versatility and positioned them as valuable
tools for serving the objectives of neuro-symbolic AI.  We provided an overview of tractable circuits in this article, 
while focusing on their foundations, their salient properties, and some of the key developments and milestones that have contributed to their current status in the field. 
There is much more to say about tractable circuits beyond what has been covered in this treatment. 
This includes the various handcrafted architectures, the learning
of circuits structures (and parameters) from data, and the compilation of these circuits from higher-level models and
other forms of knowledge.

\section*{Acknowledgements}
This work has been partially supported by grants from NSF IIS-1910317, 
ONR N00014-18-1-2561, and DARPA N66001-17-2-4032. I wish to thank Yizuo Chen, 
Haiying Huang, Jason Shen and the anonymous reviewers for providing valuable feedback.

\newpage

\bibliographystyle{plain}

\bibliography{reference}

\end{document}